\documentclass[acmtog,nonacm]{acmart}

\AtBeginDocument{%
  }

\usepackage{float} % in preamble
\usepackage{subcaption}
\usepackage{booktabs}          % For professional table lines
\usepackage{array}             % For better column control
\usepackage{multirow}          % For spanning rows (citation)
\usepackage{xcolor}
\usepackage{pifont}
\usepackage{textcomp} % Required for \textdagger
\usepackage{makecell} % optional (nicer line breaks)
\usepackage[ruled,vlined]{algorithm2e}

\newcommand{\gcmark}{{\color{green}\ding{52}}} % Green checkmark
\newcommand{\rxmark}{{\color{red}\ding{55}}}      % Red cross
\copyrightyear{2026}
\acmYear{2026}
\setcopyright{cc}
\setcctype{by}
\acmConference[SIGGRAPH Conference Papers '26]{Special Interest Group on Computer Graphics and Interactive Techniques Conference Conference Papers}{July 19--23, 2026}{Los Angeles, CA, USA}
\acmBooktitle{Special Interest Group on Computer Graphics and Interactive Techniques Conference Conference Papers (SIGGRAPH Conference Papers '26), July 19--23, 2026, Los Angeles, CA, USA}
\acmDOI{10.1145/3799902.3811052}
\acmISBN{979-8-4007-2554-8/2026/07}
\acmSubmissionID{263}

\newcommand{\TopAlignCell}[1]{%
  \parbox[t]{3cm}{\raggedright #1}%
}

\begin{document}

\title{Neural Particle Automata: Learning Self-Organizing Particle Dynamics}


\author{Hyunsoo Kim}
\authornote{Both authors contributed equally to this research.}
\affiliation{%
  \institution{KAIST}
  \city{Daejeon}
  \country{South Korea}
}
\email{khskhs@kaist.ac.kr}
\orcid{0000-0002-0404-1892}

\author{Ehsan Pajouheshgar}
\authornotemark[1]
\email{ehsan.pajouheshgar@epfl.ch}
\affiliation{%
  \institution{EPFL}
  \city{Lausanne}
  \country{Switzerland}
}

\author{Sabine Süsstrunk}
\email{sabine.susstrunk@epfl.ch}
\affiliation{%
  \institution{EPFL}
  \city{Lausanne}
  \country{Switzerland}
}

\author{Wenzel Jakob}
\email{wenzel.jakob@epfl.ch}
\affiliation{%
  \institution{EPFL}
  \city{Lausanne}
  \country{Switzerland}
}

\author{Jinah Park}
\affiliation{%
  \institution{KAIST}
  \city{Daejeon}
  \country{South Korea}
}
\email{jinahpark@kaist.ac.kr}



\begin{abstract}
We introduce \emph{Neural Particle Automata} (NPA), a Lagrangian generalization of Neural Cellular Automata (NCA) from static lattices to dynamic particle systems. 
Unlike classical Eulerian NCA where cells are fixed to pixels or voxels, NPA represent each cell as a particle with a continuous position and an internal state, both updated by a shared learnable neural rule.
This particle-based formulation yields clear individuation of cells, allows heterogeneous dynamics, and focuses computation on regions where activity is present.
However, particle systems introduce two challenges: neighborhoods are dynamic, and a naive implementation of local interactions scales quadratically with the number of particles. 
We address these issues by replacing grid-based perception with differentiable Smoothed Particle Hydrodynamics (SPH) operators, backed by memory-efficient CUDA kernels for scalable end-to-end training. 
Across tasks including morphogenesis, point-cloud classification, and particle-based texture synthesis, we show that NPA retain key NCA behaviors such as robustness and regeneration, while enabling new behaviors specific to particle systems. These results position NPA as a compact neural model for learning self-organizing particle systems.
\end{abstract}





\definecolor{pinkl}{HTML}{FB2B97}
\begin{teaserfigure}
  \centering
    \captionsetup{type=figure}
    \includegraphics[width=0.95\linewidth,trim={0 0 0 0},clip]{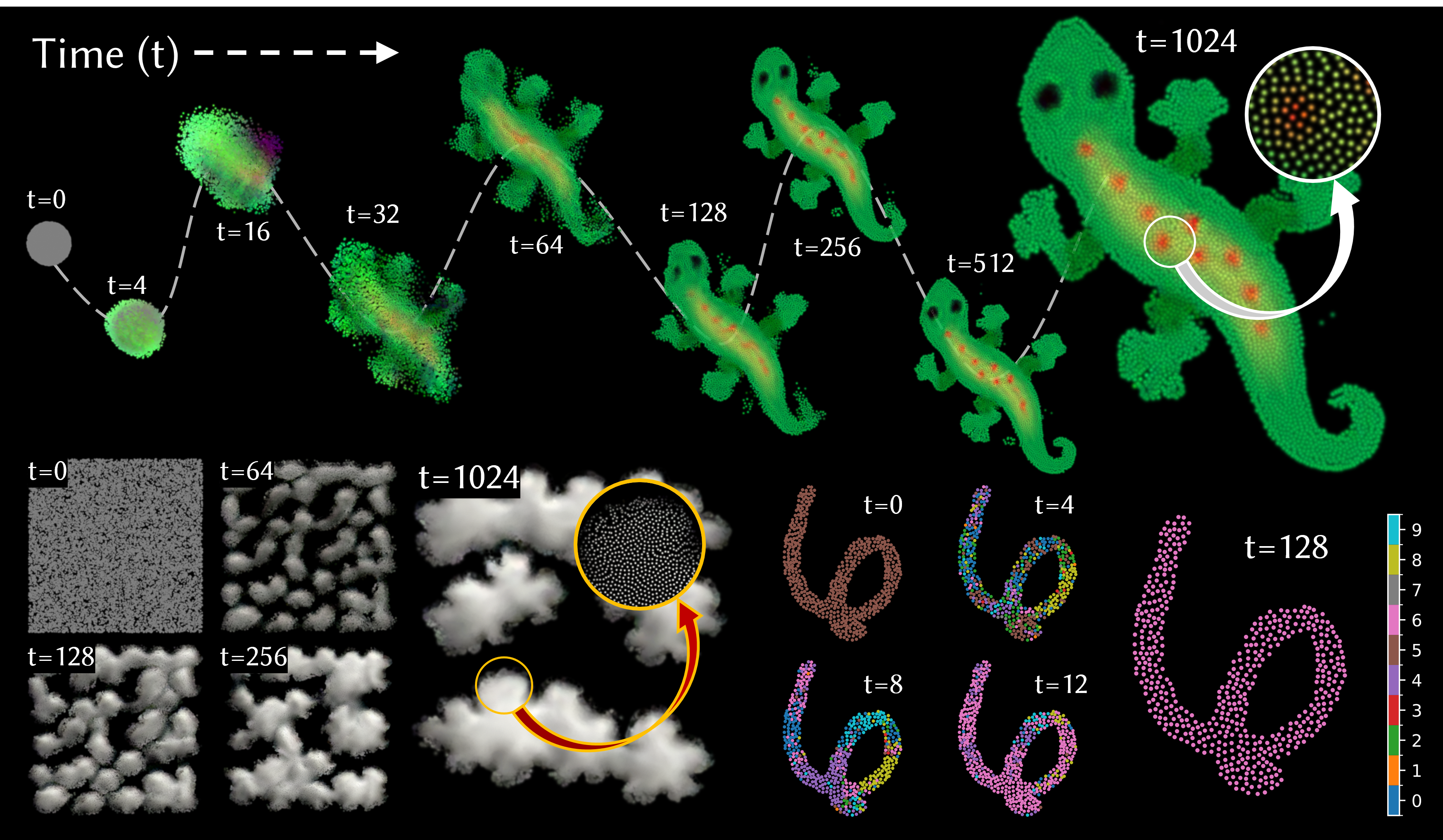}
    \captionof{figure}{
    \textbf{Neural Particle Automata} (NPA) are self-organizing particle dynamical systems driven by a shared, strictly local neural update rule. 
    We show three distinct learned rules: (top) growing a morphology from an egg-like seed, (bottom left) RGBA texture formation from a uniform seed, and (bottom right) self-classifying particle digits; numbers denote time steps. 
    Interactive web demo available at: \textcolor{pinkl}{\href{https://selforg-npa.github.io/}{\textbf{https://selforg-npa.github.io/}}}.
    }
    \label{fig:teaser}
\end{teaserfigure}

\maketitle

\section{Introduction}

Many natural phenomena exhibit self-organization: from magnetization in physical systems to morphogenesis and pigmentation in biology, and collective behaviors in bird flocks and ant colonies~\citep{selforg}. In these systems, large numbers of simple entities interact only locally, yet collectively give rise to coherent global structure and behavior. Rather than being specified by their final appearance, these patterns emerge from local interactions and self-organize into a global configuration, as exemplified by reaction-diffusion systems~\citep{turing-pattern} and flocking models such as Boids~\citep{reynolds1987flocks}. This self-organizing view offers distinct advantages over explicit, globally specified representations: it naturally extends to unbounded domains, generalizes across different resolutions and topologies due to its local nature, does not rely on global synchronization, and can progressively correct errors and recover from perturbations over time. These properties make self-organizing systems particularly attractive for interactive and real-time graphics applications.

Several computational models have been proposed to realize self-organizing representations and dynamics. 
Cellular automata~\citep{vonneumann-introca} are among the earliest such models, defined by discrete states and local interaction rules. 
Reaction-diffusion systems, originally introduced by \citet{turing-pattern}, were later adapted to computer graphics for texture synthesis~\citep{witkin1991reaction}. 
More recently, Model Synthesis~\citep{merrell2010model} and the Wave Function Collapse~\citep{gumin2016wave} have demonstrated self-organizing pattern generation in discrete domains.

Neural Cellular Automata (NCA)~\citep{mordvintsev2020growing} take this idea further by replacing hand-crafted rules with neural networks: each cell has an internal state and repeatedly updates it based on local perception, with parameters learned from data or task objectives. 
Despite their simple update rule, they can produce robust and visually rich behaviors and have been successfully applied to tasks including texture synthesis~\citep{niklasson2021self, meshnca}, modeling growth~\citep{mordvintsev2020growing, cells2pixels}, and distributed classification~\citep{randazzo2020self, kalkhof2023med}, establishing NCA as a compact and expressive framework for learning self-organizing dynamics.

Despite this progress, existing NCA models share a structural limitation: cells are pinned to a fixed lattice with static neighborhood relations. 
This Eulerian view leads to wasted computation in inactive regions and is poorly suited to heterogeneous dynamics, which requires cells to be treated as individual entities.
Extensions of NCA to irregular domains such as graphs and meshes relax the regularity of the lattice but still assume fixed connectivity. 
In parallel, self-organizing systems such as Clusters~\citep{clusters}, Particle Life~\citep{particlelife}, and Particle-Lenia~\citep{Mordvintsev_Niklasson_Randazzo_2022} operate directly on particles with local interaction, but their dynamics are hand-designed and not trained from data. 
To the best of our knowledge, there is currently no framework that combines the robustness and expressivity of NCA with the unique advantages of particle-based systems.

In this work, we introduce \emph{Neural Particle Automata} (NPA), a Lagrangian generalization of NCA from static lattices to dynamic particle systems. 
An NPA models each cell as a particle with a continuous position and an internal state, both updated by a shared, learnable rule. 
We obtain a particle-based analogue of NCA perception by replacing grid-based convolutions with mesh-free, differentiable Smoothed Particle Hydrodynamics (SPH) operators, which estimate densities, gradients, and other neighborhood features from nearby particles. 
These SPH-based features are then fed into a neural update rule that jointly adapts each particle's state and position, while preserving the strictly local, shared-rule structure of NCA.
This structure makes NPA robust, lightweight\footnote{Parameterized by a neural network with as few as ten thousand parameters.}, and well suited for real-time graphics applications.
To make this practical at scale, we implement the SPH perception using memory-efficient, CUDA-accelerated kernels that avoid naive all-pairs interactions and dramatically improve training and inference speed as well as memory usage for large particle sets.

We evaluate NPA on a diverse set of tasks in 2D and 3D. 
First, we train NPA to self-organize from simple seeds into target morphologies, with particles dynamically rearranging to form the desired shapes. 
Second, we apply NPA to distributed classification of 2D and 3D point clouds, where local interactions between particles collectively give rise to a global prediction. 
Third, we use NPA for particle-based texture synthesis, where particles self-organize to match target textures. 
Across these settings, we show that NPA retain the robustness and regenerative behavior of NCA while exploiting the sparsity and flexibility of particle-based representations.  
The real-time interactive demo and the source code are available at: \textcolor{pinkl}{\href{https://selforg-npa.github.io/}{\textbf{https://selforg-npa.github.io/}}}.

\section{Related Works}
\label{sec:related}

In this section, we provide background on NCA and discuss prior literature that motivates our approach of combining SPH operators with NCA architectures.

\subsection{Neural Cellular Automata}
NCA~\citep{mordvintsev2020growing} are neural architectures inspired by classical cellular automata, in which an iterative local update rule is learned from data. 
This yields localized computation without global aggregation and iterative updates which do not rely on global synchronization. Despite the simplicity of this local non-linear rule, repeated application over time can give rise to complex global behavior.

Since their introduction, NCA have found applications in diverse tasks including texture synthesis~\citep{niklasson2021self,dynca, meshnca, larsson2025mokume}, modeling growth and morphogenesis~\citep{mordvintsev2020growing,cells2pixels}, and distributed classification and segmentation~\citep{randazzo2020self,kalkhof2023med}. Compared to conventional neural architectures, NCA offer several advantages: parameter efficiency through simple local rules; localized processing, which makes them hardware-friendly and enables spatial scalability; robustness to perturbations, and spatiotemporal generalization, as their updates can be interpreted as discretized partial differential equation (PDE) models~\citep{noisenca}, enabling inference under varied spatial and temporal step sizes. These properties make NCA well suited for interactive and real-time graphics scenarios where compactness, locality, and robustness are desirable.

A key feature of NCA is the decoupling between the learnable update rule and the spatial perception operator. Early work used fixed Sobel and Laplacian filters on regular grids~\citep{mordvintsev2020growing,niklasson2021self}, while later variants replaced these with operators defined on graphs~\citep{grattarola2021learning}, meshes~\citep{meshnca}, and other irregular discretizations~\citep{niklasson2021self,mordvintsev2022growing,kim2025train,diffrd}. This modularity allows the same local update rule to be reused across different domains by changing only the perception operator, enabling generalization across meshes~\citep{meshnca}, from quadrilateral grids to Voronoi and hexagonal grids~\citep{niklasson2021self,mordvintsev2022growing}, and from 2D to 3D domains~\citep{kim2025train,diffrd}. 
This perspective highlights the central role of the spatial operator in neural networks, which we review in the following section.

\subsection{Spatial Operators in Neural Networks}

A \emph{spatial operator} in a neural network is any mechanism that propagates information across space by aggregating features over regions of the domain (e.g. pixels, vertices, points), before passing them to subsequent layers.
In NCA, this corresponds to the perception stage where a cell gathers information from its neighbors before applying the neural update rule. 
A wide variety of spatial operators have been proposed across architectures; \autoref{tab:spatial-operators-simple} summarizes representative examples.
These families differ in the domains they support, their degree of locality and scalability, and how well they generalize across resolutions or samplings.

\begin{table}[tb]
\centering
\caption{Comparison of spatial operators based on desired properties.
\textbf{U}: Does it work on unstructured data? \textbf{S}: Does it exploit the spatial structure in data? \textbf{DY}: Can it be effectively applied to dynamic data? \textbf{LO}: Is the influence of single operation local? \textbf{SC}: Is it scalable with increasing data resolution? \textbf{UR}: Does it generalize over unseen resolutions/samplings?
}
\label{tab:spatial-operators-simple}
\begin{tabular}{@{}rcccccc@{}}
\toprule
\textbf{Spatial Operators} & \textbf{U} & \textbf{S} & \textbf{DY} & \textbf{LO} & \textbf{SC} & \textbf{UR} \\ \midrule
Convolutions        & \rxmark & \gcmark  & \rxmark & \gcmark  & \gcmark & \rxmark \\
Graph convolutions & \rxmark & \gcmark & \gcmark & \gcmark & \gcmark & \rxmark \\
Set operators & \gcmark & \rxmark & \gcmark & \rxmark & \gcmark & \rxmark \\
Attention mechanism & \gcmark & \rxmark & \gcmark & \rxmark & \rxmark & \rxmark \\ 
Mesh differentials & \rxmark & \gcmark & \rxmark & \gcmark & \gcmark & \gcmark  \\ 
SPH operators & \gcmark & \gcmark & \gcmark & \gcmark  & \gcmark  & \gcmark  \\ \bottomrule 
\end{tabular}
\end{table}

Convolution is one of the most common operators for processing images or data on regular grids.
A large body of work extends convolution to point sets by combining learned kernels with explicit neighborhood queries~\citep{li2018pointcnn,thomas2019kpconv,wu2019pointconv}. 
While both point-based convolutions and NPA require neighborhood queries, the role of the neighborhood operator differs: point convolutions \emph{learn} the spatial kernel (and often its aggregation behavior) which can be sensitive to particle positions in the neighborhood, whereas NPA uses fixed SPH operations as sampling-robust perception mechanism.

Graph Neural Networks (GNNs) offer an alternative based on graph message passing~\citep{grattarola2021learning}, making them natural for data with explicit connectivity such as meshes or skeletons. 
However, standard message passing is topology-first and often more sensitive to the graph structure than to the underlying geometry. %, unless geometric information is explicitly encoded. 
Geometry-aware and mesh-specific extensions partly address this issue~\citep{hanocka2019meshcnn,liu2023egnn}, but they still rely on explicit, typically static connectivity, making them ill-suited for handling dynamic systems where neighborhoods change over time.

Set-based operators treat inputs as unordered point sets and aggregate features in a permutation-invariant way. 
PointNet~\citep{qi2017pointnet} employs such global set operators for spatial communication, with PointNet++~\citep{qi2017pointnet++} adding hierarchical pooling to recover some notion of locality, but permutation invariance still makes it difficult to capture oriented local geometry. 
Attention mechanisms~\citep{vaswani2017attention} provide a more expressive global operator by learning pairwise weights via key-query matching, and have been adapted to point clouds and other spatial domains~\citep{zhao2021point}. 
However, global attention is computationally expensive for high-resolution spatial data and typically relies on downsampling or windowing, while local attention variants tend to lose long-range interactions and require additional positional encodings to represent geometry faithfully.

Differential operators on meshes have recently been used as spatial operators in neural networks, with promising robustness to discretization. 
DiffusionNet~\citep{sharp2022diffusionnet} uses heat diffusion on manifolds, DeltaConv~\citep{wiersma2022deltaconv} applies gradient and divergence operators to scalar/vector features, and PoissonNet~\citep{maesumi2025poissonnet} incorporates a Poisson solver to give each layer global spatial support. 
These methods generalize well across discretizations because differential operators are robust to mesh tessellation, allowing networks to learn continuous representations rather than grid-specific features.
However, they still rely on connectivity to define the operators, which limits their applicability to particle-based systems.

Smoothed Particle Hydrodynamics (SPH)~\citep{gingold1977smoothed,lucy1977numerical} provides a mesh-free way to approximate differential operators via kernel-weighted sums over neighboring particles. 
Compared to other mesh-free schemes such as GMLS~\citep{mirzaei2012generalized}, SPH avoids matrix inversion, which simplifies implementation and improves numerical stability.
See \citet{koschier2022survey} for a comprehensive overview on SPH methods. 
Previous work that combines SPH with neural networks uses SPH within differentiable simulators, mainly for fluid modeling and control~\citep{schenck2018spnets,toshev2024jax,winchenbach2025diffsph}. 
In contrast, NPA employ SPH as a mesh-free perception operator within an NCA-style iterative local update rule, enabling diverse self-organizing behaviors beyond fluid dynamics.

SPH is a radial basis function (RBF) approach to mesh-free numerical method for PDEs.
RBF interpolation has been used with neural network as RBF networks~\citep{broomhead1988radial}, later used in deep learning~\citep{hryniowski2019deeplabnet} and neural field representation~\citep{chen2023neurbf}.
However, RBF networks utilize RBF interpolation only for point-wise nonlinearity, while NPA use SPH method for spatial message passing between nodes, allowing deeper understanding of spatial relations in data.

\section{Method}
NCA operate on a set of cells, each with an internal state $\mathbf{S}_i$ and a fixed position $\mathbf{x}_i$.
In classical NCA, $\mathbf{x}_i$ are fixed to a lattice and the neighborhood structure is given implicitly by the lattice, so perception is implemented with grid operators such as convolutions. 
In this work, we incorporate particle-based perception with differentiable SPH operators, which provide mesh-free estimates of local neighborhood features, preserving the strictly local, shared-rule structure of NCA while extending NCA on unstructured particles.

\begin{table}[b]
\centering
\caption{SPH operators used in NPA and their output dimension.}
\label{tab:sph_operators}
\resizebox{\linewidth}{!}{%
\begin{tabular}{l l c}
\toprule
\textbf{Operator} & \textbf{Formula} & \textbf{Dim.} \\
\midrule
Density &
$\displaystyle \rho_i = \sum_j m_j W_{\epsilon}(\mathbf{r}_{ji})$ &
$1$
\\[0.8em]
Smoothing &
$\displaystyle \tilde{\mathbf{S}}_i = \sum_j \frac{m_j}{\rho_j} \mathbf{S}_j W_{\epsilon}(\mathbf{r}_{ji})$ &
$C$
\\[0.8em]
Density gradient &
$\displaystyle \nabla \rho_i = \sum_j m_j W^{\nabla}_{\epsilon}(\mathbf{r}_{ji})$ &
$D$
\\
\TopAlignCell{Moment Matrix} &
$\displaystyle \mathbf{M}_i = \sum_j \frac{m_j}{\rho_j}\,\mathbf{r}_{ji}\,W^{\nabla}_{\epsilon}(\mathbf{r}_{ji})^{T}$ &
$D \times D$
\\[0.8em]
\TopAlignCell{Gradient, 0-th order \\ (difference formula)} &
$\displaystyle \nabla_{0}\mathbf{S}_i = \sum_j \frac{m_j}{\rho_j}\,(\mathbf{S}_j-\mathbf{S}_i)\,W^{\nabla}_{\epsilon}(\mathbf{r}_{ji})$ &
$C \times D$
\\[0.8em]
\TopAlignCell{Gradient, 1-st order \\ \citep{bonet1999variational}} &
$\displaystyle \nabla_{1}\mathbf{S}_i = \mathbf{M}_i^{-1} \nabla_{0}\mathbf{S}_i\,$ &
$C \times D$
\\[0.8em]
\bottomrule
\end{tabular}%
}
\end{table}

\subsection{Differentiable SPH Operators}
Smoothed Particle Hydrodynamics (SPH) is a mesh-free numerical method that represents continuous fields with discrete particles and evaluates field quantities via kernel-weighted sums over nearby samples within a \textbf{support radius} $\epsilon$. 
This kernel-weighted interpolation provides direct, local estimates of differential operators from particles without explicit connectivity.

\subsubsection{Mathematical Formulation for SPH Methods}

We denote the inputs as a set of point samples $(\mathbf{x}_i \in \mathbb{R}^D, m_i \in \mathbb{R}, \mathbf{S}_i \in \mathbb{R}^C)$, where each tuple represents the position in $D$-dimensional space, the mass, and the sampled value of a $C$-dimensional field for particle $i$, respectively.
SPH is parameterized by a support radius $\epsilon \in \mathbb{R}$ and kernel functions $W_{\epsilon}: \mathbb{R}^D \to \mathbb{R}$ and $W^{\nabla}_{\epsilon}: \mathbb{R}^D \to \mathbb{R}^D$, which take $\mathbf{r}_{ji} = \mathbf{x}_j - \mathbf{x}_i$ as input and return scalar weight or vector weight gradient, respectively. \autoref{tab:sph_operators} summarizes the SPH operators used in our differentiable framework. 
We use the Poly6 kernel $W_{\epsilon}$ for smoothing and Spiky kernel $W^{\nabla}_{\epsilon}$ for gradient, which are defined in 2D space as:
\begin{equation}
\begin{aligned}
W_{\epsilon}(\mathbf{r}) &= \frac{4}{\pi \epsilon^{8}}\left(\epsilon^{2}-\|\mathbf{r}\|^{2}\right)^{3}, 
&& 0 \le \|\mathbf{r}\| \le \epsilon,\\
W^{\nabla}_{\epsilon}(\mathbf{r}) &= \frac{10}{\pi \epsilon^{5}}\left(\epsilon-\|\mathbf{r}\|\right)^{2}\frac{\mathbf{r}}{\|\mathbf{r}\|},
&& 0 < \|\mathbf{r}\| \le \epsilon.
\end{aligned}
\label{eq:poly6_spiky}
\end{equation}
Using spiked kernels such as $W_\epsilon^\nabla$, rather than $\nabla W_{\epsilon}$, may not preserve momentum and entropy in high-accuracy simulations~\citep{dehnen2012improving}.
However, it has demonstrated better numerical stability with plausible realism in interactive applications~\citep{muller2003particle}, which our use of SPH operators is more aligned with.
Especially, $W^{\nabla}_{\epsilon}$ always gives greater magnitude of gradients as $\mathbf{r}\rightarrow 0$, which is an expected behavior as particles get closer.

\subsubsection{Backward Formulation}

We analytically derive the backward formula for each SPH operator to enable end-to-end differentiable optimization.
For example, the gradient of the loss $\mathcal{L}(\tilde{\mathbf{S}})$ with respect to a single channel sampled value $S_i$ is written as
\begin{equation}
\frac{\partial \mathcal{L}}{\partial S_i} = \sum_j \frac{\partial \mathcal{L}}{\partial \tilde{S}_j} \frac{\partial \tilde{S}_j}{\partial S_i} = \sum_j \frac{\partial \mathcal{L}}{\partial \tilde{S}_j} \frac{m_i}{\rho_i} W_\epsilon(\mathbf{r}_{ij}).
\end{equation}
\label{eq:sph-backward}
Note the backward formulation has the same neighborhood aggregation structure as the forward formulations in \autoref{tab:sph_operators}.
Thus, the backward pass can be implemented within the same computational framework as the forward pass, without additional intermediate arrays.
See Appendix for backward formulations of other operators.

The 0th-order SPH gradient $\nabla_0\mathbf{S}_i$ (\autoref{tab:sph_operators}) is a simple difference-based estimator that can be biased under irregular sampling. 
The 1st-order variant corrects this bias using the \emph{moment matrix} $\mathbf{M}_i$, which summarizes the local neighbor geometry and serves as a normalization that makes the estimate exact for locally linear fields:
\(
\nabla_1\mathbf{S}_i = \mathbf{M}_i^{-1}\nabla_0\mathbf{S}_i.
\)
In practice, however, differentiating through $\mathbf{M}_i^{-1}$ can be unstable and costly, so we use the following alternative,
\begin{equation}
\nabla \mathbf{S}_i \;=\; \nabla_0 \mathbf{S}_i \;+\; \Big(\nabla_1 \mathbf{S}_i - \nabla_0 \mathbf{S}_i\Big)\texttt{.detach()},
\end{equation}
which uses the corrected gradient for the forward pass while backpropagating as if $\nabla_0$ were used. 
This removes backpropagation through $\mathbf{M}_i^{-1}$ and improves training stability; when $\det(\mathbf{M}_i) < 10^{-3}$, we skip the correction and fall back to $\nabla_0$.

\subsubsection{CUDA Implementation}

For dynamic particle systems, neighborhoods change over time, and expressing the resulting neighborhood aggregation in connectivity matrices can lead to large, inefficient computational graphs and significant overhead. 
We implement mesh-free SPH perception using custom differentiable CUDA kernels with hash-grid acceleration to make training and inference scalable in both runtime and memory.
In contrast to differentiable SPH toolkits targeting inverse problems in fluid dynamics (e.g., JAX-SPH~\citep{toshev2024jax}, DiffSPH~\citep{winchenbach2025diffsph}), our operators target a general-purpose feature processing for neural networks.
Thus, we designed our operators with batch processing and arbitrary feature dimensionality in mind.

We bin particles into a uniform hash-grid with cell size equal to the support radius \(\epsilon\), and apply a permutation to obtain a \emph{cell-contiguous} memory layout, similar to \citet{goswami2010interactive}. 
SPH operators then scan the \(3^D\) adjacent cells and apply the true radius test before accumulating contributions. 
To balance simplicity and performance, we provide two forward kernels: (i) a vanilla thread-per-particle implementation that uses Morton hashing to improve cache locality, and (ii) a more optimized kernel that utilizes shared-memory by assigning thread blocks to grid cells (splitting crowded cells when needed) where threads cooperatively read and write consecutive neighbor-cell spans into shared memory in fixed-size strides, enabling coalesced global loads and reuse across particles in the block.
See Appendix for detailed CUDA implementation.

\subsection{Neural Particle Automata}

We propose \emph{Neural Particle Automata}, which combines SPH-based, mesh-free perception with the local, shared-rule structure of NCAs. 
Each particle computes a compact perception vector from nearby particles via SPH operators, and a small shared MLP maps this vector to state updates (and optional position updates). 
We describe the update rules for static and dynamic particles and summarize practical techniques that stabilize training.

\begin{figure}[t]
    \centering
    \includegraphics[width=\linewidth]{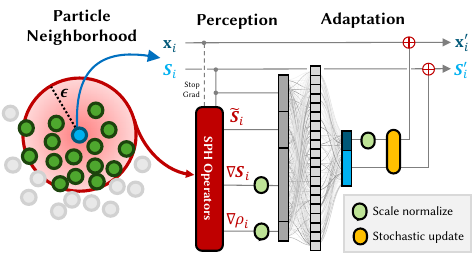}
    \caption{Single update step of Neural Particle Automata.}
    \label{fig:architecture}
\end{figure}

\subsubsection{NPA update rule}

At each step, every particle $i$ forms a perception vector from its $\epsilon$-neighborhood using SPH operators,
\begin{equation}
\mathbf{Z}_i = \big[\mathbf{S}_i,\;\tilde{\mathbf{S}}_i,\;\nabla \mathbf{S}_i,\;\nabla \rho_i\big],
\end{equation}
and applies a shared adaptation network $f_\theta$ to predict an update. 
We use a similar update scheme for both static and dynamic particle sets. 
Given the perception vector $\mathbf{Z}_i$, the adaptation network $f_\theta$, which consists of a two-layer MLP with a ReLU activation, predicts an increment
\begin{equation}
\Delta \mathbf{y}_i = f_\theta(\mathbf{Z}_i) = \mathbf{W}_2(\mathbf{W}_1\mathbf{Z}_i+\mathbf{b}_1)_+,
\end{equation}
where $\Delta \mathbf{y}_i=\Delta\mathbf{S}_i$ for \emph{static} particles and $\Delta \mathbf{y}_i=[\Delta\mathbf{x}_i;\Delta\mathbf{S}_i]$ for \emph{dynamic} particles.
Each iteration of the NPA update rules, as shown in \autoref{fig:architecture} applies an additive increment to the state (and position if needed), analogous to a single forward-Euler integration step:
\begin{align}
\mathbf{S}_i &\leftarrow \mathbf{S}_i + \delta\,\Delta \mathbf{S}_i, \\
\mathbf{x}_i &\leftarrow \mathbf{x}_i + \delta\,\Delta \mathbf{x}_i.
\end{align}
We use a stochastic update mask $\delta \sim \mathrm{Bernoulli}(p)$ with update probability $p=0.5$ following \citet{mordvintsev2020growing}.
The mask updates a random subset of particles per step, reducing reliance on global synchronization.
We study test-time behavior under different values of $p$ in \S~\ref{sec:npa_properties}.

Perception produces a compact set of local features: the particle's own state $\mathbf{S}_i$ together with neighborhood summaries $\tilde{\mathbf{S}}_i$ and $\nabla\mathbf{S}_i$, which can be interpreted as value- and derivative-like measurements of the underlying field. In particular, $(\mathbf{S}_i-\tilde{\mathbf{S}}_i)$ carries diffusion/Laplacian-like information (roughly $(\mathbf{S}_i-\tilde{\mathbf{S}}_i)\propto \epsilon^2 \Delta \mathbf{S}_i$), so while SPH formulations for Laplacian exist, we do not use an explicit SPH Laplacian in perception.
We additionally include the density gradient $\nabla\rho_i$ as a geometric cue specific to particle discretizations, where sampling density can vary across space and time. 
Because $\tilde{\mathbf{S}}$ and $\nabla\mathbf{S}$ are normalized to be density-insensitive, $\nabla\rho$ provides the missing signal of local crowding and sparsity, and directional information about where density increases or decreases.

\subsubsection{Equivariance of NPA}

SPH-based perception is inherently invariant to permutations of particle indices and to global translations by construction. 
We additionally design NPA to be equivariant to changes in sampling density and spatial scale, so the same learned rule can be applied under different particle counts and spatial length scales. 
While we do not enforce rotational equivariance in this work, it can also be incorporated, for instance by using Vector Neurons~\cite{deng2021vector} for gradient inputs in the adaptation stage.

To generalize across different sampling densities---i.e., using more or fewer particles while keeping the support radius $\epsilon$ fixed---we normalize particle masses so that the total mass remains constant (set to $1$). 
Concretely, if the particle count increases by a factor $k$, we scale each particle mass by $1/k$, which keeps the magnitude of $\nabla\rho$ consistent. 
Note that $\tilde{\mathbf{S}}$ and $\nabla\mathbf{S}$ are designed to be invariant to particle count as they are normalized by $\rho$ by definition. 

Uniformly scaling space and the support radius as $(\mathbf{x},\epsilon)\mapsto (c\mathbf{x},c\epsilon)$ represents the same configuration at a different scale and should induce consistent behavior (up to rescaling). 
Under this transformation, the scale-dependent inputs are scaled: $\nabla \mathbf{S}$ by $1/c$, and $\nabla \rho$ by $1/c^{D+1}$.
For dynamic particles, we scale the displacement $\Delta \mathbf{x}$ by $\epsilon$ for the network to learn scale-independent motion.

\subsubsection{Training NPA}

\begin{figure}[t]
    \centering
    \includegraphics[width=\linewidth]{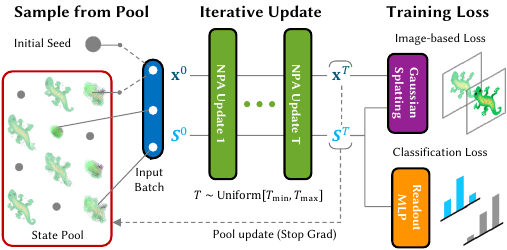}
    \caption{Training of Neural Particle Automata.}
    \label{fig:training}
\end{figure}

\autoref{fig:training} summarizes our training pipeline. 
In addition to standard NCA practices---sampling a variable number of update steps, training with a state pool for persistence, and using an overflow regularizer to keep $\mathbf{S}$ within $[-1,1]$---we introduce several techniques to stabilize the NPA training. 

Vector-valued perception terms (e.g., $\nabla\mathbf{S}$ and $\nabla\rho$) can occasionally have very large magnitudes and destabilize training. 
Inspired by logarithmic response curves in biological sensory systems, we scale each vector $\mathbf{v}$ as
\begin{equation}
\mathbf{v} \gets \log(1 + |\mathbf{v}|) \frac{\mathbf{v}}{|\mathbf{v}| + \eta},
\label{eq:log_scaling}
\end{equation}
where $\eta > 0$ is a small constant for numerical stability.

For dynamic particles, we stop gradients with respect to positions in the SPH perception, as shown in \autoref{fig:architecture}. 
Empirically, this improves stability and convergence and enables larger learning rates.
To discourage sporadic motion, we regularize the total displacement over rollout steps,
$\sum_t \|\Delta \mathbf{x}^{(t)}\|$, encouraging smooth, consistent trajectories.
For image-supervised tasks, we decode per-particle states into renderable attributes and rasterize them with Gaussian splatting to obtain color and density maps. 
Task-specific losses are then applied on these rendered outputs; we provide more details on the decoding and losses for each task in the following section.

\section{Experiments}
\label{sec:experiments}

We evaluate NPA on three representative tasks that highlight different aspects of learned self-organization: (i) \textbf{morphogenesis}, where a dynamic particle set self-organizes from a simple seed into a target shape in 2D and 3D; (ii) \textbf{particle-based texture synthesis}, where dynamic particles form stochastic texture patterns; and (iii) \textbf{point-cloud classification}, where a static point set performs distributed classification via iterative local communication. 
We also conduct a set of analyses to illustrate characteristic test-time properties of our NPA model.
Our configuration is shown in \autoref{tab:exp_config}.
See Appendix for ablation studies on the architecture and the hyperparameters.
\textbf{All figures are best viewed digitally when zoomed in.}

\begin{table}[b]
\centering
\caption{Training and model configuration for all experiments.}
\label{tab:exp_config}
\resizebox{\linewidth}{!}{%
\begin{tabular}{llcccc}
\toprule
 & \textbf{Setting}
& \makecell[c]{\textbf{Growing 2D}\\\textbf{Morphology}} & \makecell[c]{\textbf{Growing 3D}\\\textbf{Morphology}}  & \makecell[c]{\textbf{Particle}\\\textbf{Textures}} & \makecell[c]{\textbf{Point-Cloud}\\\textbf{Classification}} \\
\midrule
\multirow{6}{*}{\textbf{NPA}} 
& \#Particles ($N$)          & 4096   & 16384  & 4096  & 512 \\
& SPH radius ($\epsilon$) & 0.1    & 0.1    & 0.2   & 0.1 \\
& Seed             & \makecell[l]{Uniform ball\\(radius 0.2)} 
                             & \makecell[l]{Uniform ball\\(radius 1.0)} 
                             & \makecell[l]{Uniform square\\(side length 2.0)} 
                             & PointMNIST \\
& Channels ($C$)        & 16     & 24     & 16    & 16 \\
& MLP width              & 128    & 256    & 128   & 256 \\
& \#Parameters               & 11k    & 38k    & 11k   & 21k + 7k \\
\midrule
\multirow{7}{*}{\textbf{Training}} 
& $[T_{\min},T_{\max}]$ & [32, 96] & [24, 48] & [16, 32] & [12, 24] \\
& Batch size                 & 8      & 4      & 8     & 64 \\
& Iterations                 & 30k    & 20k    & 12k   & 50k \\
& Learning rate              & $0.0005$   & $0.003$   & $0.0001$  & $0.001$ \\
& Weight decay              & 0   & 0   & 0.1  & 0.1 \\
& Pool size                  & 512    & 16     & 512   & 1024 \\
\bottomrule
\end{tabular}%
}
\end{table}

\subsection{Morphogenesis Modeling}

Morphogenesis highlights the ability of NPA to self-organize complex structure from a simple, unstructured initial configuration. 
We initialize particles uniformly inside a ball with zero state---an "egg-like" setting in which all matter is present from the start and must be reorganized into the target form---and evolve them using a shared, trainable local rule. 
Training supervises only the \emph{rendered outcome}, allowing particles to freely rearrange as long as the rendered result matches the target. 
As a result, NPA acts as an implicit, dynamical representation of morphologies: global structure emerges through repeated local interactions rather than from an explicit parameterization or particle-to-target correspondence.

\subsubsection{2D Shapes}

\begin{figure}[]
    \centering
    \includegraphics[width=\linewidth]{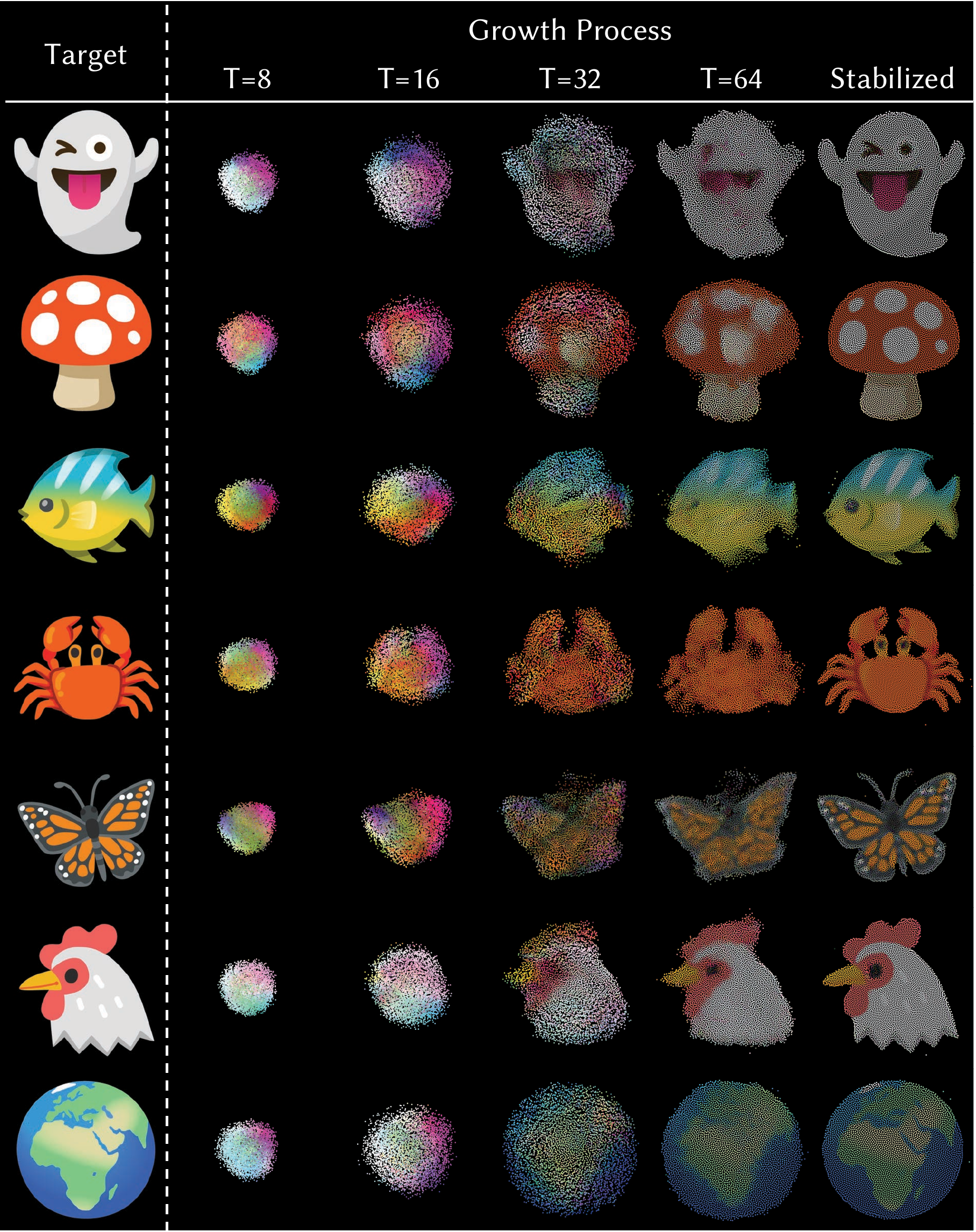}
    \caption{\textbf{Growing 2D Morphology} Starting from the same egg-like seed, learned NPA rules self-organize into different emoji targets.}
    \label{fig:growing2d}
\end{figure}

\begin{figure}[]
    \centering
    \includegraphics{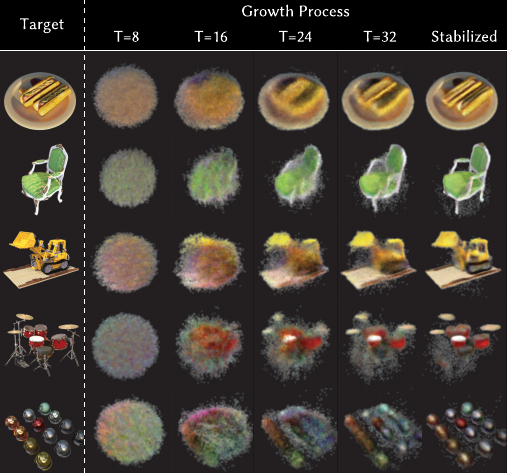}
    \caption{\textbf{Growing 3D Morphology}  A learned NPA rule grows a gaussian-splat representation from a compact seed under 3D multi-view supervision.}
    \label{fig:growing3d}
\end{figure}

We train NPA to grow colored 2D shapes from a simple seed on a dataset of $60$ emoji targets.
To define image-space supervision without correspondences, we first convert each target emoji into a colored point set by sampling roughly $4096$ points from the target image. 
We then render both the generated particle set and the target point set onto a fixed grid using isotropic Gaussian splatting, producing a color image and a density map for each.

Particle color is obtained from the last three state channels. 
We supervise both geometry and appearance using losses on the rendered density and color maps. 
To encourage stable training, we weight the color loss by a detached function of the density error so that the model first prioritizes placing mass correctly (learning shape) before refining color. 
Concretely, letting $D, C$ denote the rendered density and color, and $D^\star, C^\star$ their targets, we use a $\ell_1 + \ell_2$ loss for density $\mathcal{L}_{\text{dens}}(D,D^\star)$ and color $\mathcal{L}_{\text{col}}(C,C^\star)$, with the color term down-weighted in regions of large density mismatch (with the weight detached). Qualitative results for 2D morphogenesis are shown in \autoref{fig:growing2d}.
Compared to grid-based GrowingNCA~\citep{mordvintsev2020growing}, our particle-based NPA is more memory efficient for the same details while taking more time.
See Appendix for quantitative comparisons.

\subsubsection{3D Objects}

We additionally train NPA to self-organize into 3D objects under multi-view supervision, in the spirit of 3D Gaussian splatting. 
Each particle's position and states $(\mathbf{x}_i, \mathbf{S}_i)$ is converted into Gaussian parameters ($\mu_i, \mathbf{c}_i, \mathbf{q}_i, \mathbf{\sigma}_i, o_i$) as follows:
\begin{align*}
    \mu_i &= \mathbf{x}_i, &\text{(Position)}\\
    \mathbf{c}_i &= \mathbf{S}_i[12{:}24], &\text{(View-dependent color)}\\
    \mathbf{q}_i &= \texttt{normalize}(\mathbf{S}_i[8{:}12]), &\text{(Rotation)}\\
    \mathbf{A}_i &= A \tanh(\mathbf{S}_i[4{:}8] / A), &\text{(Anisotropy)}\\
    \mathbf{\sigma}_i &= \sigma \exp (\mathbf{A}_i[0{:}3]), &\text{(Scale)}\\
    o_i &= o \exp (\mathbf{A}_i[3]), &\text{(Opacity)}
\end{align*}
where $A, \sigma, o$ represent the maximum anisotropy, default scale, and default opacity, respectively.
$A=0.1, \sigma=0.02, o=0.15$ are used for the experiments.
Gaussians are rendered into color and alpha images using GSplat~\citep{ye2025gsplat}, and then multi-scale SSIM is used as a photometric loss between rendered and target views.
In addition to the photometric loss, we penalize anisotropic outliers with an isotropy regularizer $\mathcal{L}_{iso} = \frac{1}{N}\sum_i || A_i ||_1.$

To improve foreground/background separation, we render with a white noise background during training. 
We also employ a coarse-to-fine curriculum by initially starting with blurred target image and gradually reducing the blurring radius, allowing the model to first capture coarse structure before refining fine details.
\autoref{fig:growing3d} shows the evolution of 3D morphogenesis trained on the NeRF synthetic dataset~\citep{mildenhall2021nerf}.

\subsection{Particle-based Texture Synthesis}
\label{sec:texture}

\begin{figure}[]
    \centering
    \includegraphics[width=\linewidth]{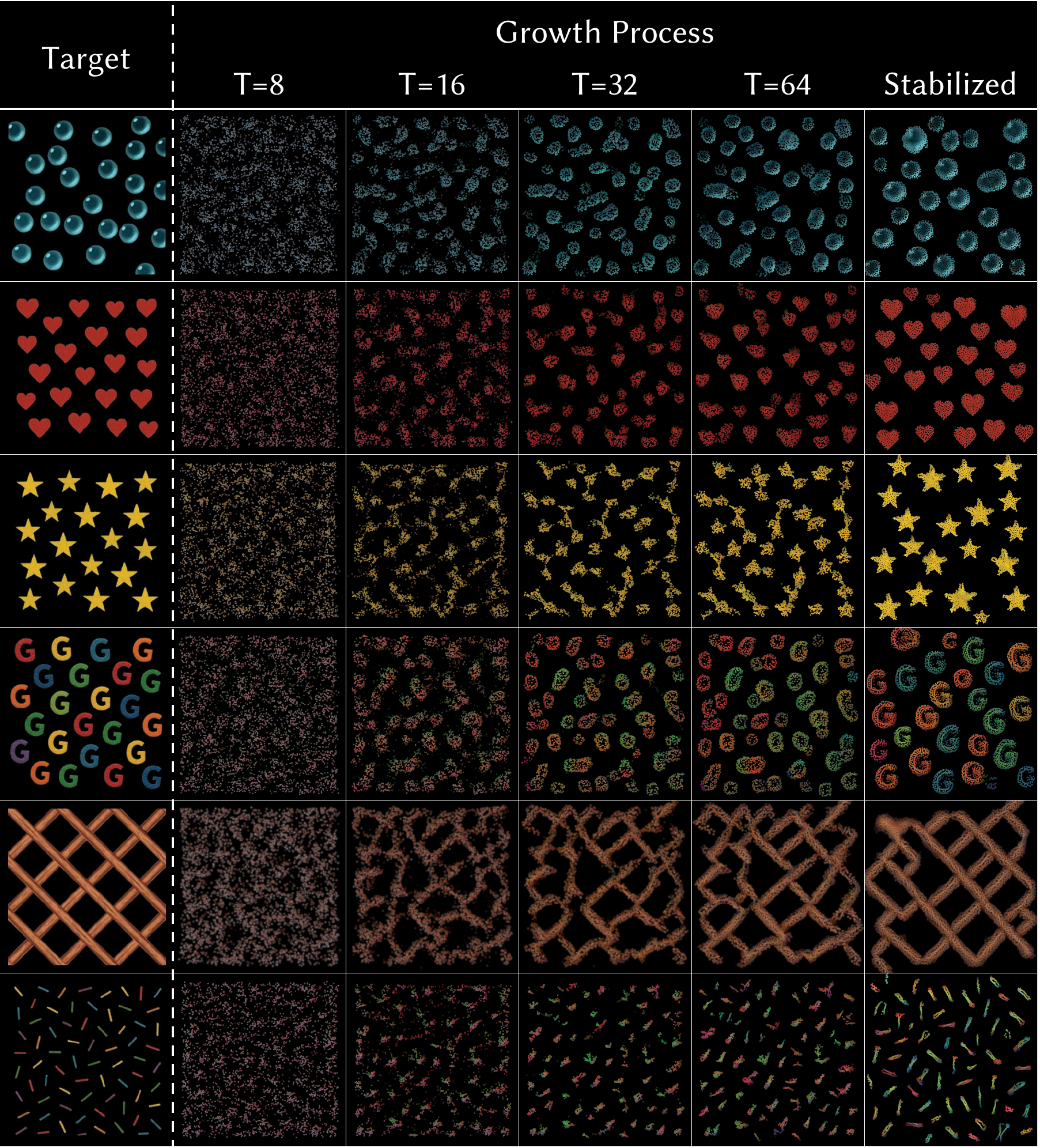}
    \caption{\textbf{Particle-based Texture Synthesis.} From a uniform square seed, a learned NPA rule self-organizes dynamic particles into textures.}
    \label{fig:texture}
\end{figure}

We next evaluate NPA on particle-based texture synthesis, shown in \autoref{fig:texture}, where the goal is to generate a target \emph{RGBA} texture with transparency by self-organizing a dynamic particle set from a uniform square initialization. 
Given particle positions and states, we use the first three channels of the state as the particle's color and render the particle set to a grid using Gaussian splatting, producing an RGB color map $\mathbf{C}$ and a scalar density map $D$. 
To define a target for supervision, we convert the RGBA texture into a point set by sampling approximately the same number of points from the target and splatting them in the same way, yielding target maps $\mathbf{C}^\star$ and $D^\star$.
Supervising both density and color allows particles to jointly match \emph{where} texture content appears (alpha) and \emph{what} it looks like (RGB).

To compare rendered outputs with targets, we use the VGG-feature optimal-transport texture loss from \citet{kolkin2019style-otloss} as adopted in \citet{dynca}. 
This loss compares multi-scale feature distributions (rather than enforcing pixel-wise alignment), which is well-suited for texture synthesis.
Since VGG features are defined on RGB images~\citep{vgg} but our rendering produces four channels, we apply the texture loss \emph{separately} to the color map and the density map: we feed $\mathbf{C}$ directly, and we repeat the density map across channels so that $D$ can be treated as a three-channel image. 

Exact density values are often less important than the \emph{support} of the density (whether particles occupy the right regions), so we apply a soft power transform
$\bar{D} = (D + \eta)^{\alpha}$, (and likewise with the target density $\bar{D^\star}$) where $\alpha=0.1$ and $\eta=10^{-6}$ is a small constant to ensure numerical stability.
This reduces sensitivity to absolute density magnitudes and empirically improves visual quality by encouraging the model to first establish the correct spatial occupancy before refining appearance.

\subsection{Self-classifying point clouds}

\begin{figure}[]
    \centering
    \includegraphics{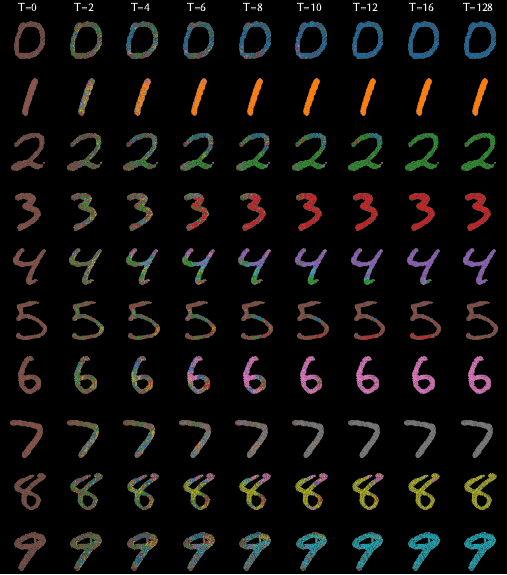}
    \caption{\textbf{Self-classifying MNIST Points.} A static point set representing MNIST digits iteratively exchanges local information until reaching a global consensus for classification. }
    \label{fig:mnist-evolution}
\end{figure}

\begin{figure}[]
    \centering
    \includegraphics{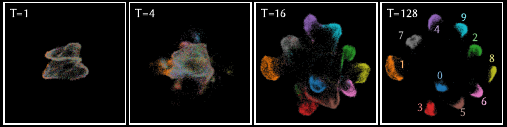}
    \includegraphics{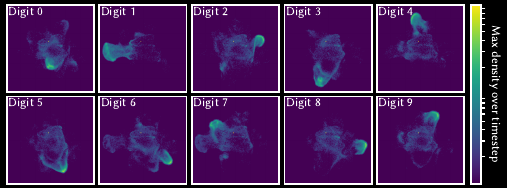}
    \caption{Internal state visualization of self-classifying Point MNIST. (Top) Evolution of internal states by digits on reduced 2D space. (Bottom) Maximum density of internal states over the total timesteps by each digit. }
    \label{fig:mnist-internal}
\end{figure}

We evaluate NPA on point cloud classification tasks to study whether global semantic decisions can emerge from purely local particle interactions. 
We create a PointMNIST dataset by sampling 512 uniform points from each MNIST image, in range $[0, 1]^2.$ 
We first upsample the image to $224^2$, and randomly sample a point from each pixel with value greater than 0.5. The number of points is then reduced to 512 using farthest point sampling.

Following \citet{randazzo2020self}, we replace the cross entropy loss with an $L_2$ loss on the one-hot target vectors. 
Empirically this stabilizes the long-term dynamics by preventing blow-ups. Our architecture achieved 98.42\% classification accuracy on the test set. 
\autoref{fig:mnist-evolution} illustrates the evolution of correctly classified examples for each digit, where every particle reaches consensus using only local operations.  
\autoref{fig:mnist-internal} shows the internal state evolution of 200 digits after dimension reduction with UMAP~\citep{mcinnes2018umap-software} and the \textit{footprint} of internal states for each digit on the reduced space.
The iterative nature of NPA lets us visualize how NPA understands the shape of digits and forms a global consensus in detail.

\subsection{NPA Properties}
\label{sec:npa_properties}

\begin{figure}[]
    \centering
    \includegraphics[width=\linewidth]{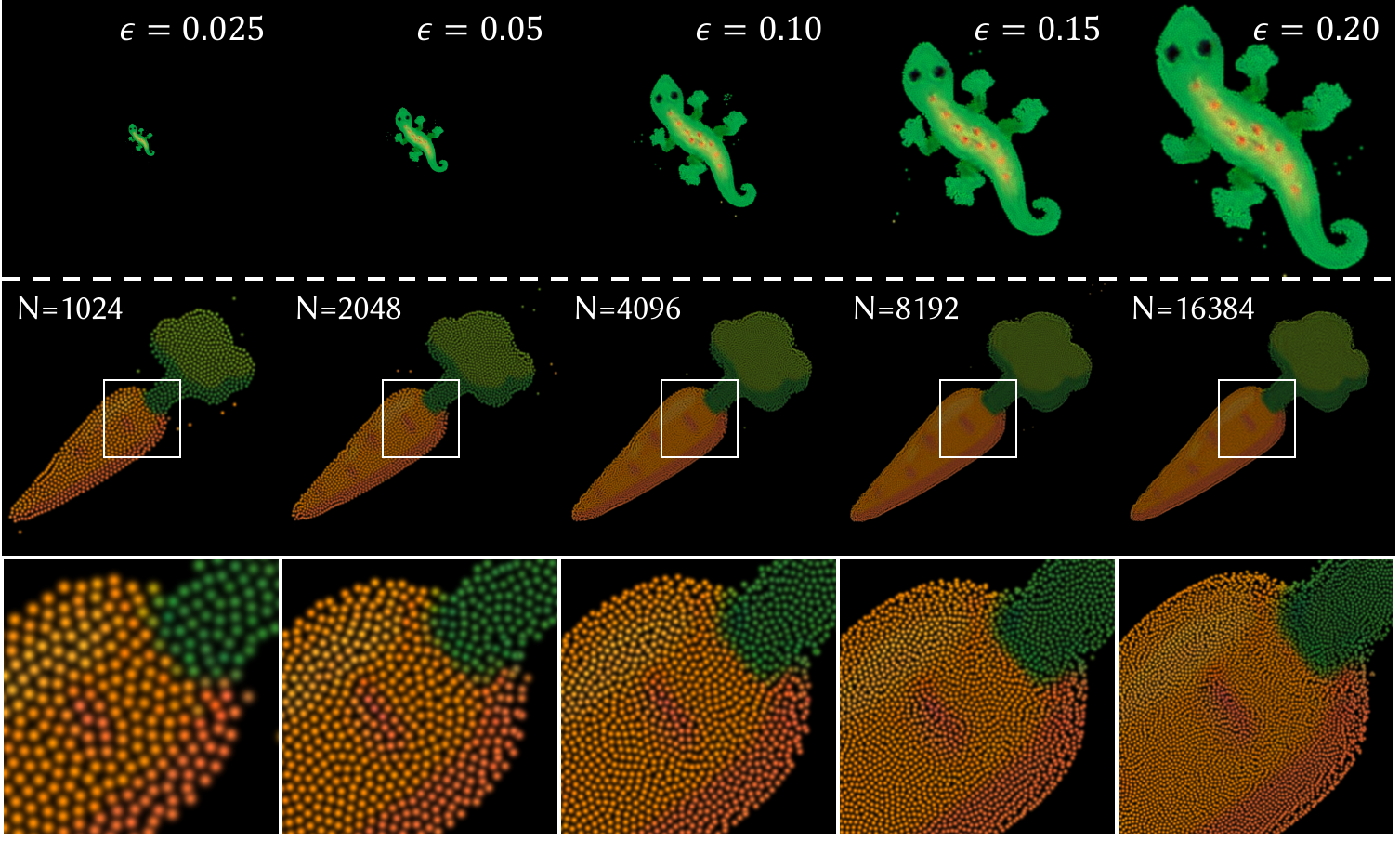}
    \caption{\textbf{Robustness to Discretizations.} Varying $\epsilon$ and $N$ at test time.}
    \label{fig:eps_N}
\end{figure}

\begin{figure}[]
    \centering
    \includegraphics[width=\linewidth]{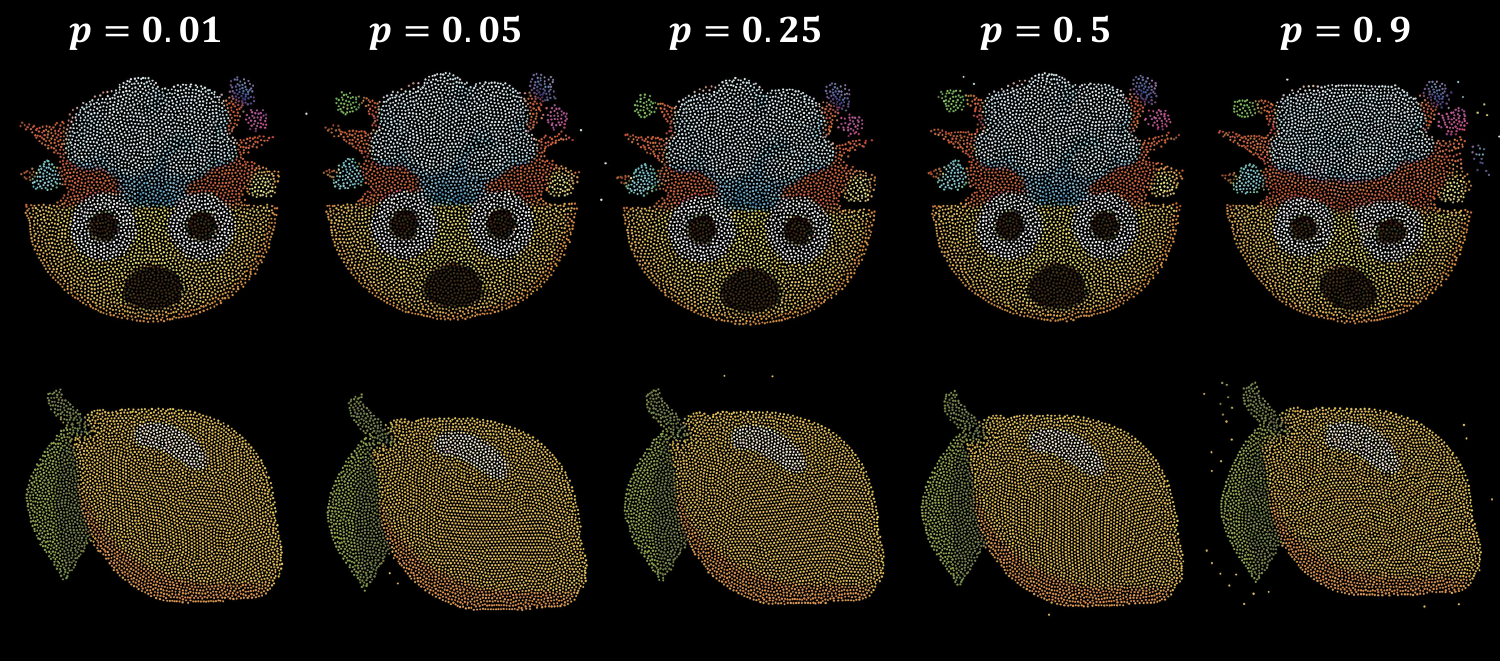}
    \caption{NPA remains stable for different stochastic update probabilities $p$.}
    \label{fig:stoch_update}
\end{figure}

NPA models have several interesting test-time properties.
\autoref{fig:eps_N} probes robustness to discretization by varying the SPH support radius $\epsilon$ and the number of particles $N$ at inference time for a model trained with $(N{=}4096,\epsilon{=}0.1)$. 
When changing $\epsilon$, we rescale the initial seed accordingly; by design, this preserves the NPA dynamics up to numerical precision (i.e., float round-off), and we observe stable behavior across a broad range of $\epsilon$.  When changing $N$, we do not have a formal equivariance guarantee, but empirically the dynamics degrade gracefully at lower particle counts and typically improve as $N$ increases, where SPH neighborhood estimates become more accurate. Zoomed-in views highlight how sampling density affects local coverage and particle spacing. Additionally, \autoref{fig:stoch_update} shows that NPA is robust to varying the update probability $p$ at inference time, despite all models being trained with $p=0.5$. In the limit $p\to 0$, the stochastic updates approach a Poisson process, yielding effectively asynchronous dynamics with no global synchronization.

\begin{figure}[]
    \centering
    \includegraphics[width=\linewidth]{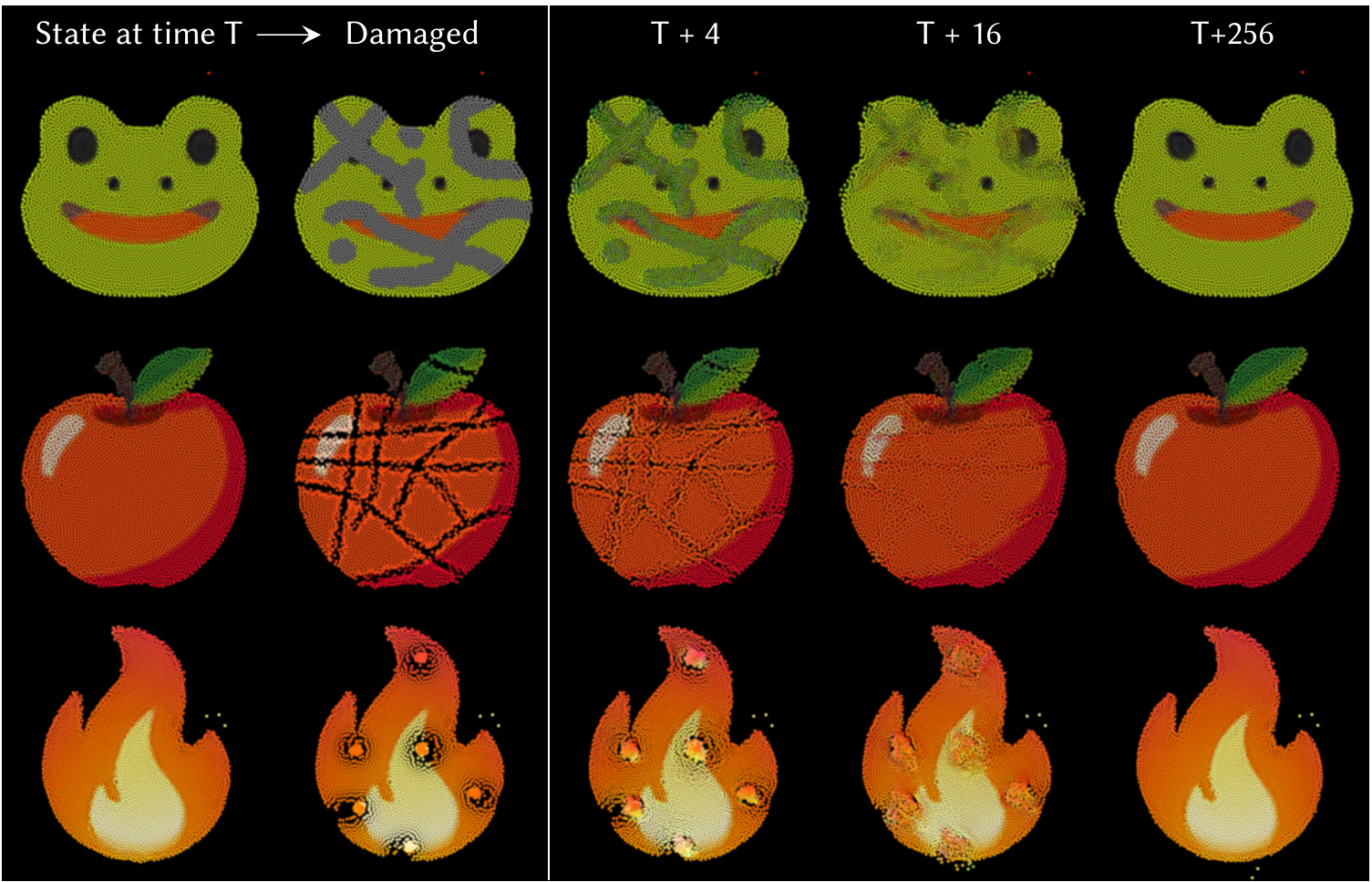}
    \caption{\textbf{Regeneration.} Continued application of the learned local rule repairs damage and recovers from state erasure, cuts, and clumping.}
    \label{fig:regen}
\end{figure}

\autoref{fig:regen} illustrates the regenerative behavior of NPA under three types of perturbations: (i) locally zeroing particle states, (ii) cut-like disruptions, and (iii) clumping particles by pulling them toward a point. In all cases, continued application of the learned local rule repairs the damage and recovers the target structure. To encourage such regeneration in the 2D morphogenesis and texture settings, we apply a simple training-time disturbance: whenever we sample an element from the pool, we pick a random particle and set the states of all particles within its $\epsilon$-neighborhood to zero. This local perturbation regularizes the dynamics toward self-correction; notably, the other two distortions are never seen during training, yet the resulting models still recover from them at test time.

\begin{figure}[]
    \centering
    \includegraphics[width=\linewidth]{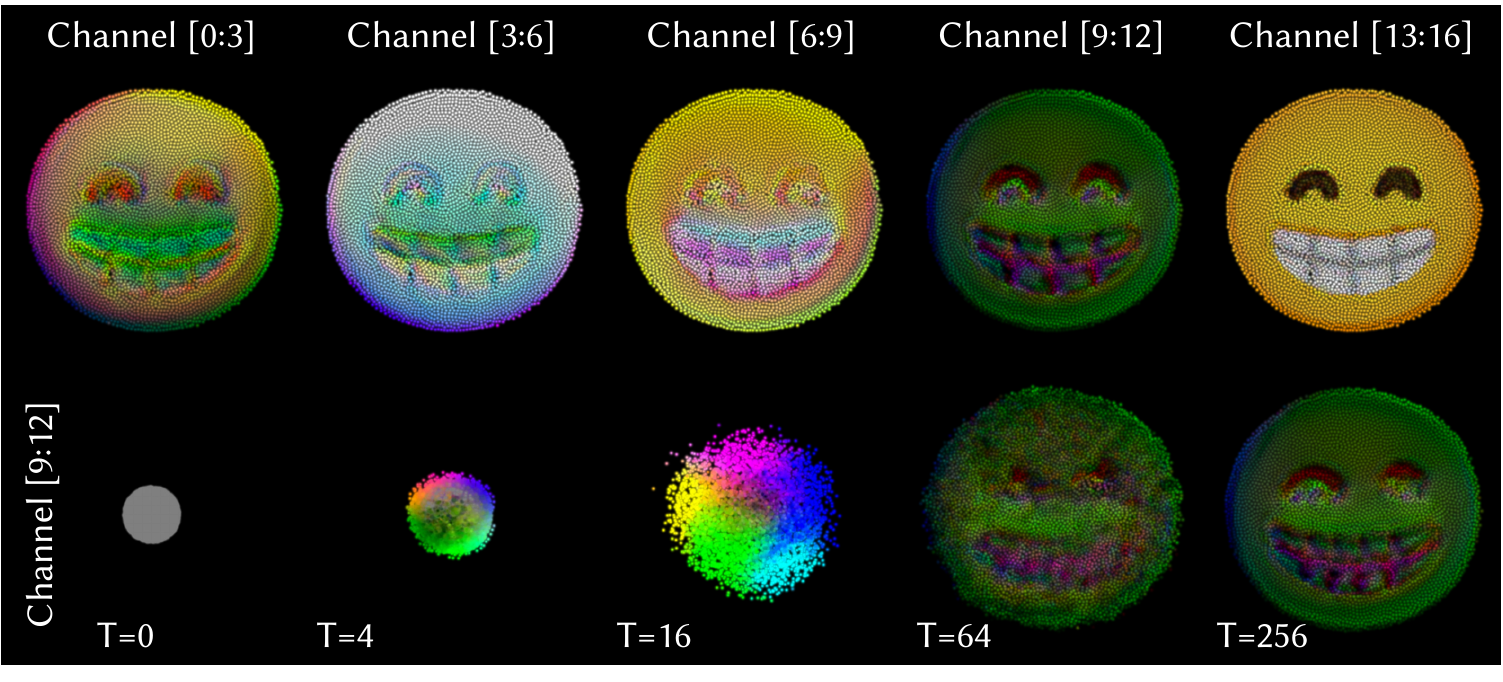}
    \caption{\textbf{Hidden Channels.} Internal states form a spatially structured representation, guiding the growth and  helping with long-term stability.}
    \label{fig:hidden}
\end{figure}

\begin{figure}[]
    \centering
    \includegraphics[width=\linewidth]{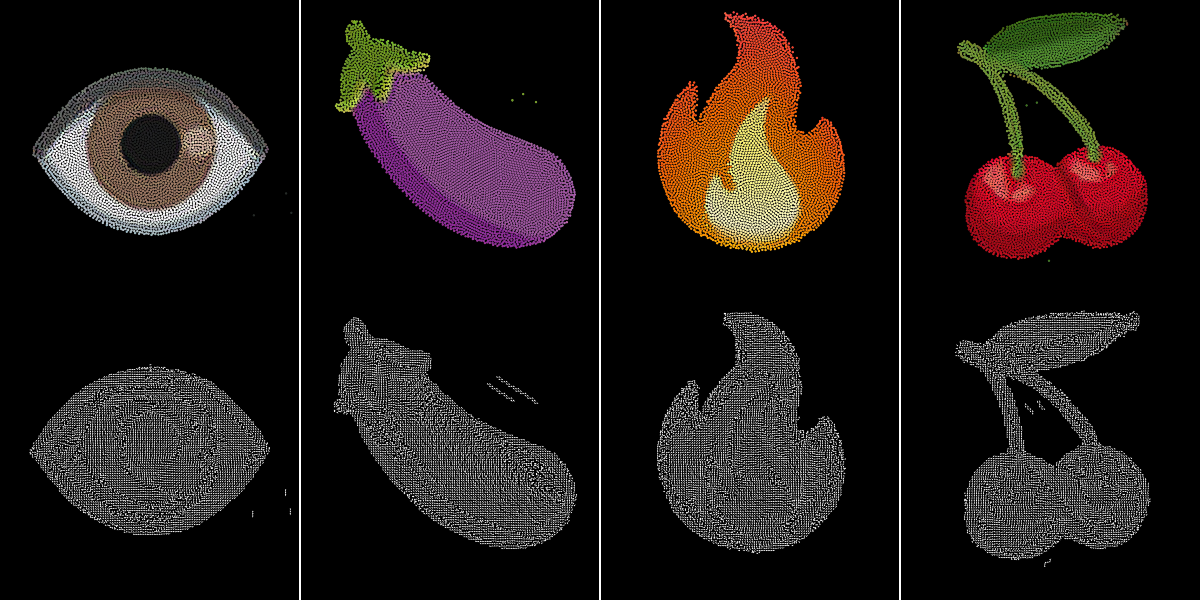}
    \caption{\textbf{Persistent Internal Flows.} Even after the target is reached, particles typically keep moving within the shape; the resulting vortex-like vector field indicates stable circulating trajectories under the learned dynamics.}
    \label{fig:vector_field}
\end{figure}

To better understand what the learned rule encodes, \autoref{fig:hidden} visualizes the hidden state channels by rendering channel groups as RGB. The channels consistently organize into spatially meaningful patterns, suggesting that they act as internal representations that help particles infer their role within the emerging morphology. In particular, channels $[9{:}12]$ often exhibit a rainbow-like structure during early stages of growth, consistent with a directional signal that guides differentiation and expansion from the seed. We also observe that, even after the target shape is reached, particles typically continue moving within the object, forming vortex-like trajectories as shown in \autoref{fig:vector_field}. Because the image-based objective does not enforce particle correspondences, such persistent flows naturally emerge as stable solutions of the learned dynamics. We speculate they may serve as a distributed memory that helps maintain the morphology over long periods.

\begin{figure}[]
    \centering
    \includegraphics[width=\linewidth]{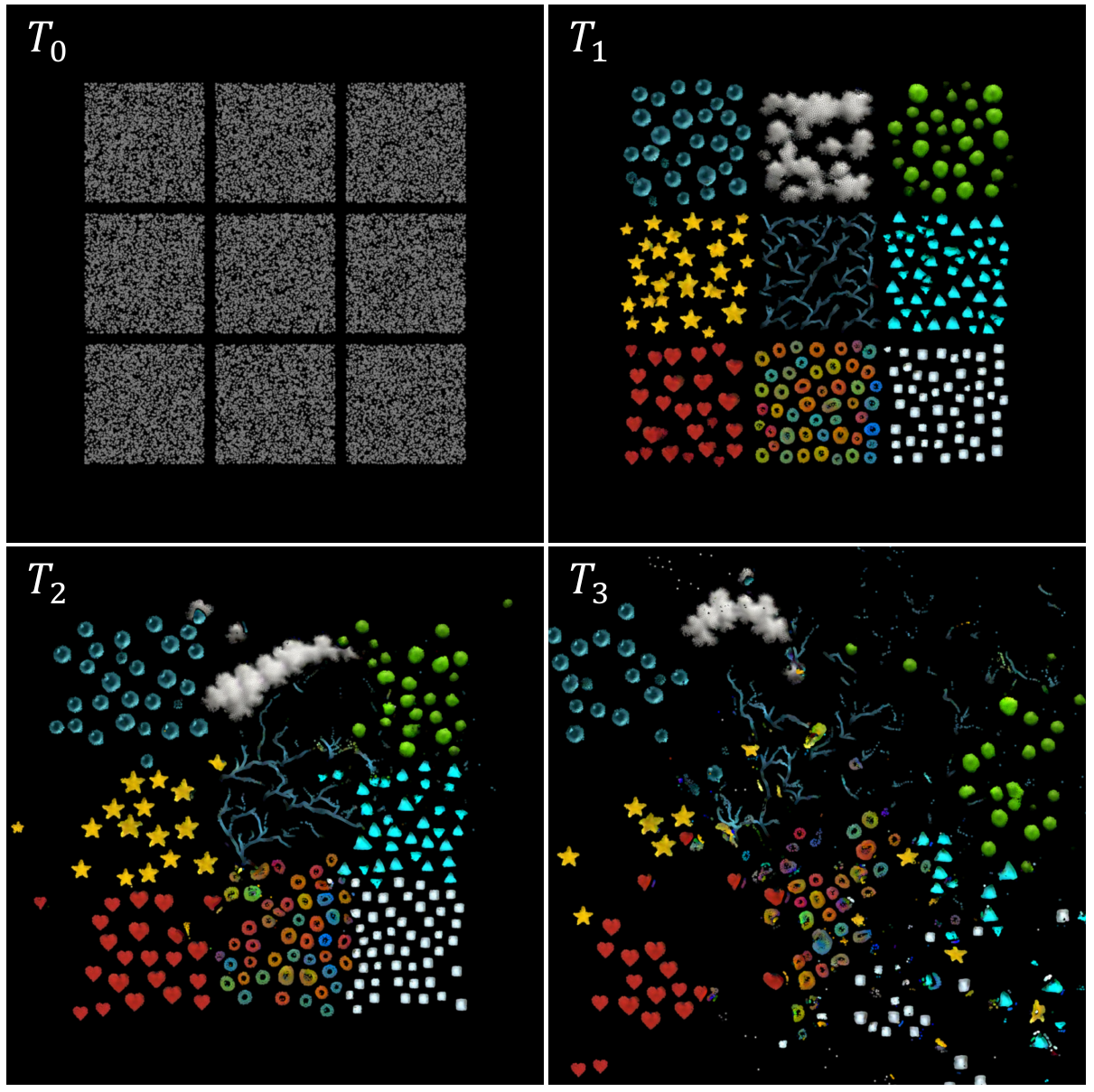}
    \caption{Interaction of multiple independently trained NPA models.}
    \label{fig:species}
\end{figure}

Finally, \autoref{fig:species} demonstrates that multiple independently trained NPA "species" can be composed within the same simulation. Each particle follows its own learned rule, and as groups move and interpenetrate they become part of one another's neighborhoods and therefore interact, despite never being co-trained. We observe a range of qualitative outcomes---from near-independence to cooperation and disruption---highlighting NPA as a candidate compact building block for multi-agent, self-organizing particle simulations.

\section{Conclusion}
\label{sec:conclusion}
We have presented Neural Particle Automata (NPA), a framework that generalizes NCA from fixed Eulerian grids to dynamic Lagrangian particle systems.  
By replacing grid-based convolutions with differentiable SPH operators and implementing them via custom CUDA kernels, we enable end-to-end training of self-organizing dynamics in continuous space. We demonstrated that NPA can perform a variety of self-organizing tasks, including morphogenesis modeling, particle-based texture synthesis, and self-classifying point clouds.  
We also showed that NPA model brings additional benefits such as perturbation robustness, controllable feature scale, generalization across particle resolutions, and support for multi-species interactions.

\paragraph{Limitations and future works.}
Several challenges emerged during our experiments. First, NPA cannot merge or split particles; consequently, when the target region is small, particles become denser and SPH operations slow down. This fixed particle count also limits the model's ability to capture fine geometric and texture details: regions that require higher resolution cannot recruit additional particles, while over-sampled regions cannot shed them. 
Second, NPA's gradient perceptions assume global Euclidean basis, limiting the model's generalization over rotations.
Finally, the learning dynamics are sensitive to hyperparameters. 
These limitations suggest interesting future works, such as learning particle merging and splitting dynamics, rotation-invariant/equivariant architectures, and searching the optimal configurations.

\paragraph{Applications.}
Built upon mesh-free and local interactions, NPA have a wide range of applications.
Because NPA directly process dynamic point data without meshing or conversion, the model can be used in tasks involving dynamic particles such as optimizing initial conditions for fluid simulations~\cite{dehnen2012improving}, and point-cloud denoising.
NPA's local perception mechanism can be easily extended beyond Euclidean geometry, such as texture synthesis on manifolds~\cite{kim2025train}.
Additionally, As a trainable self-organizing system, NPA can learn a desired behavior of distributed dynamic systems, such as display systems made of drones or crowds.

\begin{acks}
This work was supported by Institute for Information \& communications Technology Promotion(IITP) grant funded by the Korea government(MSIT) (No.00223446, Development of object-oriented synthetic data generation and evaluation methods)
\end{acks}

\bibliographystyle{ACM-Reference-Format}
\bibliography{ref}

\appendix

\section{Backward Derivations of SPH Operators}
\label{sec:sph_backward_appendix}

\paragraph{Notation.}
We denote reverse-mode (incoming) gradients by
$\bar{y} := \frac{\partial \mathcal{L}}{\partial y}$.
Particle positions are $\mathbf{x}_i \in \mathbb{R}^D$, densities are $\rho_i \in \mathbb{R}$, and states are $\mathbf{S}_i \in \mathbb{R}^C$.
We define pairwise offsets
$\mathbf{r}_{ij} := \mathbf{x}_i - \mathbf{x}_j$,
distance $d_{ij} := \|\mathbf{r}_{ij}\|$,
and neighborhood $\mathcal{N}(i):=\{j \mid d_{ij} < \epsilon\}$.
All kernels and their derivatives are identically zero for $d_{ij}\ge \epsilon$.
At $d_{ij}=0$, direction-dependent terms are set to zero in practice to avoid singularities.

\subsection{Kernels}
\label{sec:sph_backward_kernels}

We use the Poly6 kernel $W_\epsilon$ and the Spiky gradient kernel $W^{\nabla}_{\epsilon}$:
\begin{equation}
\begin{aligned}
W_{\epsilon}(\mathbf{r})
&=
Z_{\epsilon}\left(\epsilon^{2}-\|\mathbf{r}\|^{2}\right)^{3},
\\
W^{\nabla}_{\epsilon}(\mathbf{r})
&=
Z^{\nabla}_{\epsilon}\left(\epsilon-\|\mathbf{r}\|\right)^{2}\frac{\mathbf{r}}{\|\mathbf{r}\|}.
\end{aligned}
\label{eq:appendix_poly6_spiky}
\end{equation}

where $Z_{\epsilon}$ denotes the Poly6 normalization constant and $Z^{\nabla}_{\epsilon}$ denotes the Spiky-gradient normalization constant. In $2$D, we use
$Z_{\epsilon}=\frac{4}{\pi\epsilon^{8}}$ and $Z^{\nabla}_{\epsilon}=\frac{10}{\pi\epsilon^{5}}$,
while in $3$D we use
$Z_{\epsilon}=\frac{315}{64\pi\epsilon^{9}}$ and $Z^{\nabla}_{\epsilon}=\frac{15}{\pi\epsilon^{6}}$.

The backward pass requires (i) the spatial derivative of $W_\epsilon$, and (ii) the Jacobian of $W^{\nabla}_{\epsilon}$.
For $0<d=\|\mathbf{r}\|<\epsilon$, the Poly6 gradient is
\begin{equation}
\nabla_{\mathbf{r}} W_\epsilon(\mathbf{r})
=
-6Z_{\epsilon}\left(\epsilon^{2}-\|\mathbf{r}\|^{2}\right)^{2}\mathbf{r}.
\label{eq:appendix_poly6_grad}
\end{equation}
% Let $\mathbf{g}(\mathbf{r}) := W^{\nabla}_{\epsilon}(\mathbf{r})\in\mathbb{R}^D$ and define its Jacobian
% $\mathbf{H}_\epsilon(\mathbf{r}):=\frac{\partial \mathbf{g}(\mathbf{r})}{\partial \mathbf{r}}\in\mathbb{R}^{D\times D}$.

Let 
$\mathbf{H}_\epsilon(\mathbf{r}):=\frac{\partial W^{\nabla}_{\epsilon}}{\partial \mathbf{r}}\in\mathbb{R}^{D\times D}$ be the Jacobian of the Spiky gradient kernel.
For $0<d=\|\mathbf{r}\|<\epsilon$,
\begin{equation}
\mathbf{H}_\epsilon(\mathbf{r})
=
3Z^{\nabla}_{\epsilon} \left(
\frac{(\epsilon-d)^2}{d}\mathbf{I}
-\frac{\epsilon^{2}-d^{2}}{d^{3}}\mathbf{r}\mathbf{r}^{\top}
\right),
\label{eq:appendix_spiky_jacobian}
\end{equation}
where $\mathbf{I}$ is the $D\times D$ identity matrix. 

\subsection{Backward Derivations}
\label{sec:sph_backward_derivations}

\paragraph{\textbf{(1) Density}}
\begin{equation}
\rho_i = \sum_{j\in \mathcal{N}(i)} m_j\, W_\epsilon(\mathbf{r}_{ij}).
\label{eq:appendix_density_fwd}
\end{equation}
The position gradient is
\begin{equation}
\frac{\partial \mathcal{L}}{\partial \mathbf{x}_i}
=
\sum_{j\in\mathcal{N}(i)}
\left(
m_j\,\bar{\rho}_i
+
m_i\,\bar{\rho}_j
\right)\,
\nabla_{\mathbf{r}} W_\epsilon(\mathbf{r}_{ij}).
\label{eq:appendix_density_bwd_x}
\end{equation}

\paragraph{\textbf{(2) Smoothing (blur)}}
\begin{equation}
\tilde{\mathbf{S}}_i
=
\sum_{j\in \mathcal{N}(i)} \frac{m_j}{\rho_j}\,\mathbf{S}_j\,W_\epsilon(\mathbf{r}_{ij}).
\label{eq:appendix_blur_fwd}
\end{equation}
The gradients w.r.t.\ $\mathbf{S}_i$ and $\rho_i$ are
\begin{align}
\frac{\partial \mathcal{L}}{\partial \mathbf{S}_i}
&=
\frac{m_i}{\rho_i}
\sum_{j\in\mathcal{N}(i)}
W_\epsilon(\mathbf{r}_{ij})\,
\bar{\tilde{\mathbf{S}}}_j,
\label{eq:appendix_blur_bwd_S}
\\
\frac{\partial \mathcal{L}}{\partial \rho_i}
&=
-\frac{m_i}{\rho_i^{2}}
\sum_{j\in\mathcal{N}(i)}
W_\epsilon(\mathbf{r}_{ij})\,
\bar{\tilde{\mathbf{S}}}_j^{\top}\mathbf{S}_i.
\label{eq:appendix_blur_bwd_rho}
\end{align}
The position gradient is
\begin{equation}
\frac{\partial \mathcal{L}}{\partial \mathbf{x}_i}
=
\sum_{j\in\mathcal{N}(i)}
\Bigg[
\bar{\tilde{\mathbf{S}}}_i^{\top}\!\left(\frac{m_j}{\rho_j}\mathbf{S}_j\right)
+
\bar{\tilde{\mathbf{S}}}_j^{\top}\!\left(\frac{m_i}{\rho_i}\mathbf{S}_i\right)
\Bigg]\,
\nabla_{\mathbf{r}} W_\epsilon(\mathbf{r}_{ij}).
\label{eq:appendix_blur_bwd_x}
\end{equation}

\paragraph{\textbf{(3) Gradient (0-th order difference form)}}
We write the operator as
\begin{equation}
\mathbf{G}_i := \nabla^0\mathbf{S}_i
=
\sum_{j\in\mathcal{N}(i)}
\frac{m_j}{\rho_j}\,
(\mathbf{S}_i-\mathbf{S}_j)\otimes W^{\nabla}_{\epsilon}(\mathbf{r}_{ij})
\in\mathbb{R}^{C\times D}.
\label{eq:appendix_grad0_fwd}
\end{equation}
The gradient w.r.t.\ $\mathbf{S}_i$ is
\begin{equation}
\frac{\partial \mathcal{L}}{\partial \mathbf{S}_i}
=
\sum_{j\in\mathcal{N}(i)}
\left(
\frac{m_j}{\rho_j}\bar{\mathbf{G}}_i
+
\frac{m_i}{\rho_i}\bar{\mathbf{G}}_j
\right)\,
W^{\nabla}_{\epsilon}(\mathbf{r}_{ij}),
\label{eq:appendix_grad0_bwd_S}
\end{equation}
where the matrix--vector multiplication produces a $C$-vector.
The density gradient is
\begin{equation}
\frac{\partial \mathcal{L}}{\partial \rho_i}
=
-\frac{m_i}{\rho_i^{2}}
\sum_{j\in\mathcal{N}(i)}
(\mathbf{S}_i-\mathbf{S}_j)^{\top}
\bar{\mathbf{G}}_j\,W^{\nabla}_{\epsilon}(\mathbf{r}_{ij}).
\label{eq:appendix_grad0_bwd_rho}
\end{equation}
The position gradient is obtained by backpropagating through $W^{\nabla}_{\epsilon}$:
\begin{equation}
\frac{\partial \mathcal{L}}{\partial \mathbf{x}_i}
=
\sum_{j\in\mathcal{N}(i)}
\mathbf{H}_\epsilon(\mathbf{r}_{ij})^{\top}
\left(
\frac{m_j}{\rho_j}\bar{\mathbf{G}}_i
+
\frac{m_i}{\rho_i}\bar{\mathbf{G}}_j
\right)^{\top}
(\mathbf{S}_i-\mathbf{S}_j)
.
\label{eq:appendix_grad0_bwd_x}
\end{equation}

\paragraph{\textbf{(4) Density gradient}}
\begin{equation}
\nabla \rho_i = \sum_{j\in\mathcal{N}(i)} m_j\,W^{\nabla}_{\epsilon}(\mathbf{r}_{ji})
\in\mathbb{R}^{D}.
\label{eq:appendix_densgrad_fwd}
\end{equation}
Let $\bar{\mathbf{g}}_i := \frac{\partial \mathcal{L}}{\partial (\nabla\rho_i)}\in\mathbb{R}^D$.
Then
\begin{equation}
\frac{\partial \mathcal{L}}{\partial \mathbf{x}_i}
=
\sum_{j\in\mathcal{N}(i)}
\mathbf{H}_\epsilon(\mathbf{r}_{ij})^{\top}
\left(
m_i\,\bar{\mathbf{g}}_j
-
m_j\,\bar{\mathbf{g}}_i
\right).
\label{eq:appendix_densgrad_bwd_x}
\end{equation}

\paragraph{\textbf{(5) Moment matrix}}
\begin{equation}
\mathbf{M}_i
=
\sum_{j\in\mathcal{N}(i)}
\frac{m_j}{\rho_j}\,\mathbf{r}_{ji}\,W^{\nabla}_{\epsilon}(\mathbf{r}_{ji})^{\top}
\in\mathbb{R}^{D\times D}.
\label{eq:appendix_moment_fwd}
\end{equation}
Let $\bar{\mathbf{M}}_i := \frac{\partial \mathcal{L}}{\partial \mathbf{M}_i}\in\mathbb{R}^{D\times D}$
and $\langle \mathbf{A},\mathbf{B}\rangle_F := \mathrm{tr}(\mathbf{A}^{\top}\mathbf{B})$.
The density gradient is
\begin{equation}
\frac{\partial \mathcal{L}}{\partial \rho_i}
=
-\frac{m_i}{\rho_i^{2}}
\sum_{j\in\mathcal{N}(i)}
\left\langle
\bar{\mathbf{M}}_j,\;
\mathbf{r}_{ij}\,W^{\nabla}_{\epsilon}(\mathbf{r}_{ij})^{\top}
\right\rangle_F.
\label{eq:appendix_moment_bwd_rho}
\end{equation}
For the position gradient, define the pairwise coefficient matrix
\begin{equation}
\mathbf{C}_{ij}
:=
\frac{m_j}{\rho_j}\bar{\mathbf{M}}_i
+
\frac{m_i}{\rho_i}\bar{\mathbf{M}}_j
\in\mathbb{R}^{D\times D}.
\label{eq:appendix_Cij_def}
\end{equation}
Then
\begin{equation}
\frac{\partial \mathcal{L}}{\partial \mathbf{x}_i}
=
\sum_{j\in\mathcal{N}(i)}
\left[
\mathbf{C}_{ij}\,W^{\nabla}_{\epsilon}(\mathbf{r}_{ij})
+
\mathbf{H}_\epsilon(\mathbf{r}_{ij})^{\top}\left(\mathbf{C}_{ij}^{\top}\mathbf{r}_{ij}\right)
\right].
\label{eq:appendix_moment_bwd_x}
\end{equation}

\section{CUDA Implementation of SPH Operators}
\label{sec:cuda_forward_text}

We implement all forward SPH operators using a uniform hash grid and custom CUDA kernels.
Rather than storing an explicit adjacency list (which is variable-sized and expensive for dynamic particle sets), we build a compact spatial index that lets each particle enumerate candidate neighbors by scanning the small set of adjacent grid cells and filtering by the true support radius.

We first \emph{bin} particles into grid cells of size equal to the SPH support radius.
This binning produces a \emph{cell-contiguous} layout in memory: particles belonging to the same cell occupy a single contiguous interval.
Forward kernels then iterate over the \(3^D\) neighbor cells (e.g., \(3\times 3\) in 2D, \(3\times 3\times 3\) in 3D), load candidate particles from those cell intervals, and accumulate operator-specific contributions.

We provide implementations of two different approaches for parallelizing the computation and neighborhood traversal:
\begin{itemize}
    \item \textbf{Particle-Centric (Morton hashing):} One CUDA thread per particle; neighborhoods are streamed directly from global memory.
    We use Morton (Z-order) hashing so spatially nearby cells tend to be close in memory after binning, improving cache locality when each thread independently scans the \(3^D\) neighboring cells.

    \item \textbf{Grid-Centric (Row-major hashing + shared memory):} We assign CUDA blocks to grid cells (splitting high-occupancy cells across multiple blocks as needed).
    We use row-major hashing to preserve contiguity along the \(x\)-dimension, enabling threads to cooperatively load neighbor particle data from global memory into shared memory as \emph{consecutive} contiguous spans, then reuse the staged data across all particles in the block.
\end{itemize}

\subsection{Hash-Grid Binning and Data Structures}
\label{sec:hashgrid_binning_text}
For simplicity we assume all particles have an equal mass that does not change over time.
Given an input batch of particles $(\mathbf{x} \in \mathbb{R}^{B\times N \times D}, \mathbf{S} \in \mathbb{R}^{B\times N \times C})$, where $B$ is the batch size and $N$ is the number of particles in a batch, we construct:
\begin{itemize}
    \item \textbf{Cell counts} \(\texttt{cellCount}[b, h]\): number of particles that fall into cell \(h\) in batch \(b\).
    \item \textbf{Cell offsets} \(\texttt{cellOffset}[b, h]\): cumulative sum over counts, giving the start index of particles for cell \(h\) in the binned particle arrays.
    \item \textbf{Permutation} \(\texttt{perm}[b, i]\): the destination index of particle \(b, i\) in the binned layout.
    \item \textbf{Binned arrays} \(\texttt{xBin}, \texttt{SBin}\): particle positions and features permuted such that each cell occupies one contiguous range ordered with the cell index.
\end{itemize}
In the grid-centric variant, we additionally build:
\begin{itemize}
    \item \textbf{Block descriptors} \(\texttt{BlockInfo}\): a compact list that maps each CUDA block to a particular cell (and optionally an offset range within that cell for load balancing when cells are very full).
\end{itemize}

\subsection{Particle-Centric Approach}
\label{sec:vanilla_forward_text}

We assign one CUDA thread per particle in the binned arrays.
Each thread computes the output of the chosen SPH operator for its particle by iterating over neighbor candidates.

\paragraph{Neighborhood traversal.}
For a particle \(i\) a CUDA thread:
\begin{enumerate}
    \item Computes \(i\)'s cell and its hash \(h\).
    \item Enumerates the \(3^D\) adjacent cells around \(h\).
    \item For each neighbor cell, obtains the particle 
    
    range \([\texttt{cellOffset}[h_n], \texttt{cellOffset}[h_n+1])\).
    \item Iterates through that contiguous range in global memory and applies a radius check (\(\|\mathbf{x}_j-\mathbf{x}_i\|<\epsilon\)) before accumulating operator contributions.
\end{enumerate}

\paragraph{Core design points.}
\begin{itemize}
    \item \textbf{No shared memory.} This keeps the kernel simple and avoids block-level synchronization.
    \item \textbf{Morton hashing for cache efficiency.} Since neighbor scanning reads global memory directly, we rely on the binned Morton layout to improve cache hit rates and reduce memory divergence.
    \item \textbf{Stable work pattern.} Each particle does the same conceptual work (scan neighbor cells, then radius filter), with variable neighbor counts handled naturally by loops.
\end{itemize}

\subsection{Shared-Memory Forward Kernel (GridBased)}
\label{sec:gridbased_forward_text}

We assign each CUDA block to a specific grid cell.
If a cell contains many particles, we split it into multiple blocks to improve load balancing across blocks. We store the mapping between blocks and cells in \texttt{BlockInfo} data structure. 

Within a CUDA block:
\begin{itemize}
    \item We first cooperatively load the block's \emph{local particles} (those in the assigned cell chunk and any required attributes such as state or density) into shared memory.
    \item We then iterate over the neighbor cells, where all threads in the block perform cooperative, coalesced reads of neighbor particle data into shared memory.
\end{itemize}

The grid-centric kernel is designed so that neighbor particles are fetched in a small number of \emph{contiguous} global-memory spans:
\begin{itemize}
    \item \textbf{In 2D:} the block reads 3 consecutive spans corresponding to the \emph{top}, \emph{middle}, and \emph{bottom} neighbor rows.
    Each row span covers the 3 adjacent \(x\)-cells, so each row can often be fetched as one contiguous interval.
    \item \textbf{In 3D:} the block reads 9 consecutive spans corresponding to all \((\Delta y,\Delta z)\in\{-1,0,1\}^2\) neighbor rows.
    Again, each span covers the 3 adjacent \(x\)-cells.
\end{itemize}
We use \textbf{row-major hashing} in this variant specifically so that, for a fixed neighbor row, the three \(x\)-neighbor cells map to consecutive cell hashes and therefore to a contiguous particle interval after binning.

Neighbor rows may contain more particles than the available shared memory.
We therefore process each neighbor row in \emph{chunks}:
\begin{itemize}
    \item Threads cooperatively load a fixed-size chunk (STRIDE) of neighbor particle data into shared memory.
    \item All threads then compute interactions between their assigned local particles and the staged neighbor chunk.
    \item We repeat until the entire neighbor span has been processed.
\end{itemize}
This preserves coalesced loads, maximizes reuse of staged neighbor data, and keeps shared memory bounded.

\paragraph{Core design points.}
\begin{itemize}
    \item \textbf{Reuse local data.} Local particles stay in shared memory while the kernel streams over neighbor chunks, reducing repeated global reads.
    \item \textbf{Coalesced global access.} Neighbor particles are read as contiguous spans (3 spans in 2D, 9 spans in 3D), which reduces memory transaction overhead relative to per-thread random access.
    \item \textbf{Robust to variable cell occupancy.} Cells with many particles are split across multiple blocks and empty cells generate no blocks. 
\end{itemize}

\subsection{Comparison with Prior Work}
% \rev{
% Our implementation shares some similarities with the work of \citet{goswami2010interactive}. Their method can be seen as a hybrid between our particle-centric and GridBased approaches. For example, in their method each thread processes a single particle and they use Morton hashing which is similar to our particle-centric approach, while thread cooperate to load neighboring particles into shared memory and use a data structure similar to \(\texttt{BlockInfo}\) to assign threads to particles which is similar to our GridBased approach. The advantage of our GridBased approach is that we use row major hashing so that we can read chunks of 3 neighbor cells at once and in a coalesced manner which helps a lot with performance especially since each particle carries a $C$ dimensional state compared to their case for fluid simulation where each particle carries limited number of features such as position and velocity. Also in Our grid-centric approach  each thread could proccess multiple local and neighbor particles using an strided implementation}

Our implementation shares architectural similarities with the work of \citet{goswami2010interactive}, 
whose method can be viewed as a hybrid of our particle-centric and grid-centric approaches.
Like our particle-centric variant, each of their threads is responsible for a single particle 
and they employ Morton (Z-order) hashing for spatial indexing.
At the same time, threads cooperatively load neighboring particles into shared memory and 
a compact block descriptor (analogous to our \texttt{BlockInfo}) assigns thread blocks to 
spatial regions, which parallels our grid-centric design.

Our grid-centric approach differs in two key respects.
First, we use row-major hashing so that, for a given neighbor row, 
the three adjacent $x$-cells map to consecutive cell hashes and can therefore 
be fetched as a single contiguous span in a coalesced read.
This is particularly beneficial in our setting, where each particle carries 
a $C$-dimensional feature state, substantially more data per particle than 
the position-and-velocity attributes typical of fluid simulations such as theirs.
Second, our strided loop design allows each thread to process multiple local 
and neighbor particles, decoupling the thread count from the cell occupancy 
and improving utilization when cells vary in size.

% \rev{
% Additionally their method include a bunch of tricks such as caching the SPH kernels by pre-computing them and storing them in CUDA constant memory, and using a double-buffer approach to increase GPU utilization and decrease the idle time for memory read operations. We've tried integrating these ideas into our implementation however, Unfortunately their kernel caching leads to degraded performance on modern GPU hardware especially when the SPH kernels are cheap to compute (such as our smoothing and spiky kernels) because threads access pattern to this constant memory block is scattered so the read from constant memory is slower than actually computing the kernels. Their double-buffer idea slightly degrades the performance of our kernels when integrated but could prove beneficial if one tries to scale our experiments to millions of particles and more dense neighborhoods. 
% }

\citet{goswami2010interactive} additionally employ two optimization techniques:
(i)~pre-computing SPH kernel values for discretized distances and storing them 
in CUDA constant memory as lookup tables, and
(ii)~an interleaved load-compute pattern in which each thread issues global memory 
loads for the next batch of neighbor particles while simultaneously computing 
interactions against the previously loaded batch, thereby overlapping memory 
latency with arithmetic work.

We evaluated both ideas in our implementation.
The kernel lookup table degrades performance on modern hardware because 
threads within a warp evaluate the SPH kernel at different distances, 
producing divergent addresses into the constant memory array.
Constant memory is optimized for uniform (broadcast) access; 
under divergent access the reads serialize, 
making the lookup slower than directly computing our smoothing and spiky kernels, 
which require only a few floating-point operations each.
The interleaved load-compute pattern requires double-buffering the neighbor 
data in shared memory, roughly doubling the shared memory footprint 
of the neighbor-side buffers.
In our experiments this increased register pressure and reduced occupancy, 
yielding a slight net slowdown for the particle counts and neighborhood 
densities we tested.
The technique may nevertheless prove beneficial at larger scales 
with denser neighborhoods, where the compute-to-load ratio per chunk 
is high enough to fully hide the memory latency.
\section{Evaluations}
\label{sec:evaluations_appendix}
% \rev{
% In this section, we provide a qualitative comparison with the GrowingNCA model, as well as comparing the performance of our cuda-based implementation of NPA forward pass with the forward pass of a typical convolutional NCA. 
% We use an NVIDIA 4090 TI GPU for the performance benchmark. We also provide an ablation study on the key architectural choices in the design of SPH perception.
% }

We provide (i) a qualitative comparison to GrowingNCA on a shared morphogenesis task, (ii) a forward-pass scaling benchmark comparing our CUDA SPH perception against a convolutional NCA baseline, and (iii) an ablation study isolating key design choices in SPH perception and training stabilization, as well as the choice of perception radius $\epsilon$. All benchmarks are run on an NVIDIA GeForce RTX 4090 GPU.

\begin{figure}[]
    \centering
    \includegraphics[width=\linewidth]{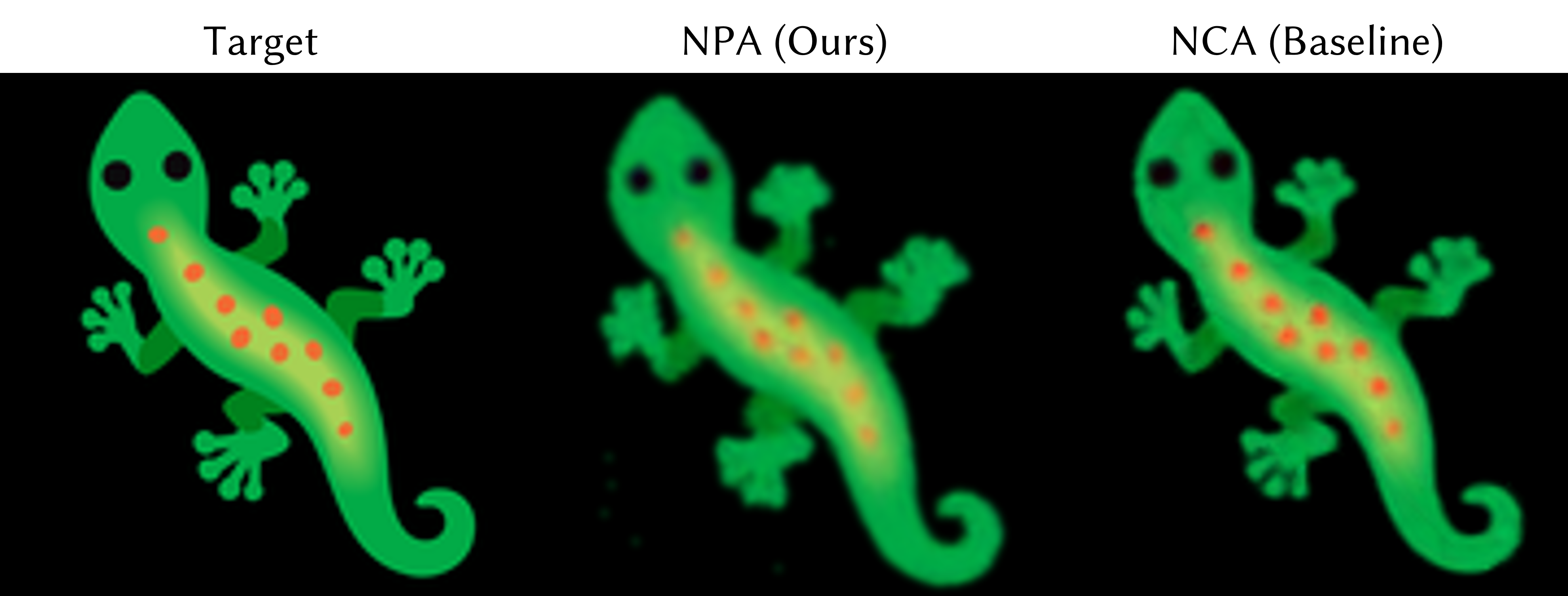}
    \caption{Qualitative comparison on morphogenesis: GrowingNCA (grid) vs.\ NPA (particles) for the same target and similar active-cell/particle budget.}
    \label{fig:comparison}
\end{figure}

\subsection{Qualitative Comparison with GrowingNCA}
% \rev{
% As there is no baseline model in the literature for learning self-organizing particle dynamics, we choose to compare our model to an NCA baseline as it is the most natural choice and since our model was a Lagrangian generalization of the Eulerian NCA model. We compare both models on a shared task of growing a morphology from a simple seed. While in NPA all the cells are present from the initial state, in GrowingNCA model the initial state only contains a single living cell and the number of living cells changes throughout the evolution of the state. We choose the NCA grid size in a way such that the total number of living cells are roughly the same for NCA, and our NPA model. For the lizard morphology to get roughly 4096 living cells, this translates to roughly a $120\times120$ NCA grid size. Both model share the same configuration such as batch size, $[T_{min}, T_{max}]$, pool size, number of channels and hidden neurons in the MLP. For the GrowingNCA model we use a learning rate of $0.001$ which is higher than the learning rate of NPA $0.0005$ as this is the common learning rate used for NCA training. 
% The NCA training takes roughly 14 minutes per each 10k iteration and uses 11.4 GB of GPU memory, while NPA training takes around 29 minutes per 10k training iterations and uses 6.2 GB of GPU RAM. 
% Figure ~\ref{fig:comparison} shows the results of our qualitative comparison. NPA and NCA reach a similar quality level. While NPA training is slower than NCA training, it uses less GPU memory. 
% }

As there is no established baseline for \emph{trainable} self-organizing \emph{particle} dynamics, we compare NPA to the most closely related neural-automata baseline: GrowingNCA.
We evaluate both methods on the same task of growing a target morphology from a simple seed. NPA starts with a fixed set of particles present from the beginning, whereas GrowingNCA starts from a single live cell and grows a variable number of active cells over time.

To make the comparison as fair as possible, we choose the GrowingNCA grid resolution so that the final number of active cells is comparable to our particle count ($N{=}4096$). For the lizard target, this corresponds to approximately a $120{\times}120$ grid.
Both models use the same training setup (batch size, $[T_{\min},T_{\max}]$, pool size, channel count, and MLP width). We use the standard GrowingNCA learning rate of $10^{-3}$, while NPA uses $5{\times}10^{-4}$.
On this setup, GrowingNCA training takes $\sim$14 minutes per 10k iterations and uses 11.4\,GB of GPU memory, while NPA takes $\sim$29 minutes per 10k iterations and uses 6.2\,GB.
\autoref{fig:comparison} shows that both methods reach comparable visual quality; NPA is slower per iteration but notably more memory efficient in this regime.

\begin{figure*}[]
    \centering
    \includegraphics[width=0.95\linewidth]{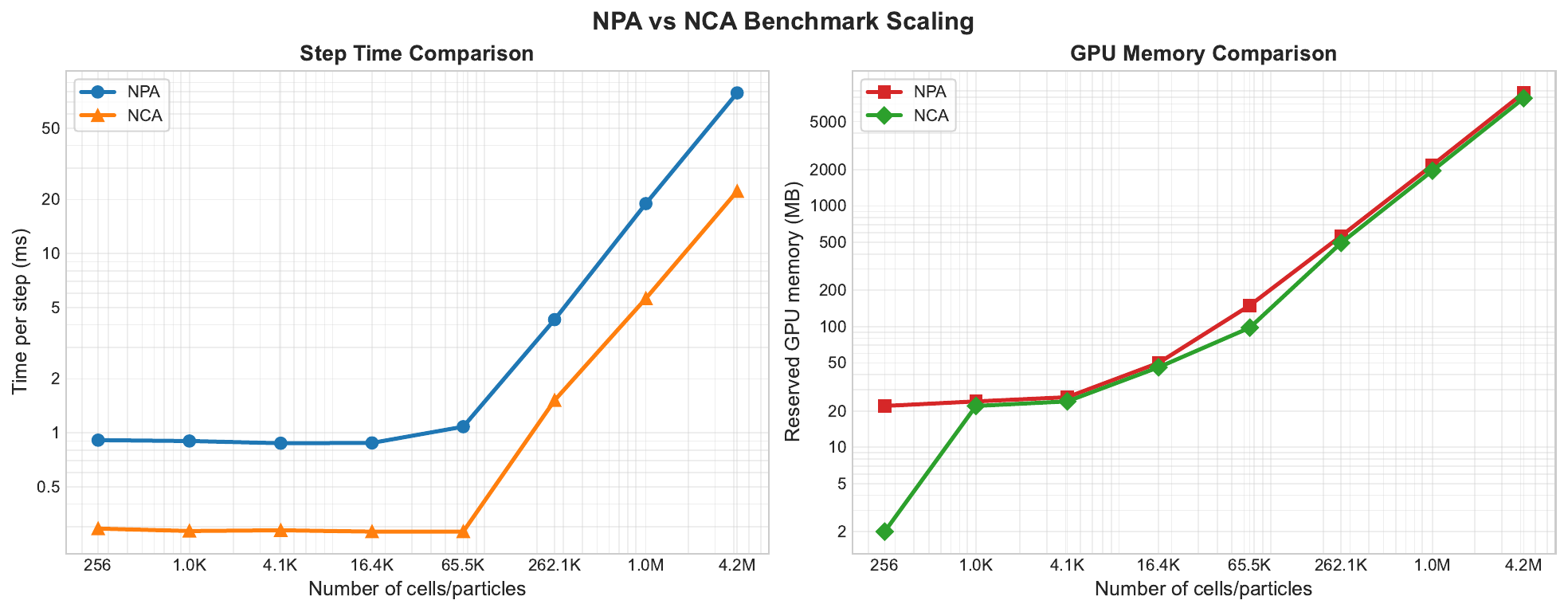}
    % \vspace{-8pt}
    \caption{Forward-step scaling on RTX 4090: step time (left) and reserved GPU memory (right) vs.\ number of particles/cells for NPA and convolutional NCA.}
    % \vspace{-10pt}
    \label{fig:performance}
\end{figure*}

\subsection{Performance Comparison with NCA}
% \rev{
% In this section we compare the inference performance of NCA and NPA. More precisely, we uniformly spread different number of particles $N$ in range $[-1.6,1.6]^2$ and set $\epsilon \approx 0.2 \sqrt{\frac{1024}{N}}$ so that the number of particles in the neighborhood stays roughly constant where each particle has around $13$ neighboring particles. We compare this to an NCA running on different grid sizes where the total number of cells is equal to the number of pixels in the grid. Note that this is a fair comparison since the NCA perception radius remains constant regardless of the grid size (a 3x3 neighborhood around each cell). For different number of cells we evaluate the time for a single forward iteration of both models as well as the required GPU memory. Figure~\ref{fig:performance} shows the results. The results show the scaling law for our cuda implementation and demonstrate that our kernels scale similarly to highly optimized implementations of convolution operators. The results show that both memory and time stay constant in low cell counts (the regime where the GPU is under-utilized) and once we reach full GPU utilization both NPA and NCA memory and time scale linearly with the number of cells. The y-offset in the time figure shows that NPA forward pass is proportionally slower than NCA due to two things: first the number of neighbor particles $13$ for NPA vs $9$ for NCA and second the fact that SPH operations are more complicated and require more floating point operations. }

We next compare inference-time scaling by measuring the wall-clock time and reserved GPU memory for a \emph{single} forward update step.
For NPA, we uniformly distribute $N$ particles in $[-1.6,1.6]^2$ and set $\epsilon \approx 0.2\sqrt{1024/N}$ to keep the expected neighbor count approximately constant (about 13 neighbors per particle).
For NCA, we vary the grid size so that the number of cells matches $N$, noting that the convolutional perception radius remains fixed (a $3{\times}3$ neighborhood, i.e., 9 neighbors).

\autoref{fig:performance} shows that both approaches exhibit a constant-time regime at small sizes (GPU under-utilization) and transition to roughly linear scaling once the GPU is saturated.
NPA has a higher constant factor in step time, consistent with (i) a slightly larger neighborhood (13 vs.\ 9) and (ii) the higher arithmetic intensity of SPH perception compared to a fixed $3{\times}3$ convolution.

\subsection{Ablation Study}
% \rev{
% We conduct an ablation study on the design of our SPH perception for the NPA model which includes three key architectural choices: the stop-grad on particle position, log normalization of gradients, omitting the SPH gradient terms from the perception, to show the necessity of all components. 
% For the ablation we use the morphology growing experiment and choose the lizard pattern as the target. 
% Figure~\ref{fig:perception_ablation} shows the training loss throughout the training for different ablations of our NPA model. As shown in the figure the directional information (state and density gradients in the SPH operators) are necessary for the model to be able to learn anything and without it the model doesn't show sign of learning. While the two other ablations show signs of learning, the training is very unstable and can diverge. Without the stop-grad on position the model diverges pretty quickly. While the log-gradient normalization seems less critical and the model shows signs of learning and partial stability, the training loss is higher and also after we flush the NPA pool at iteration 10k this variant also becomes unstable and diverges. 
% }

\begin{figure}[]
    \centering
    \includegraphics[width=\linewidth]{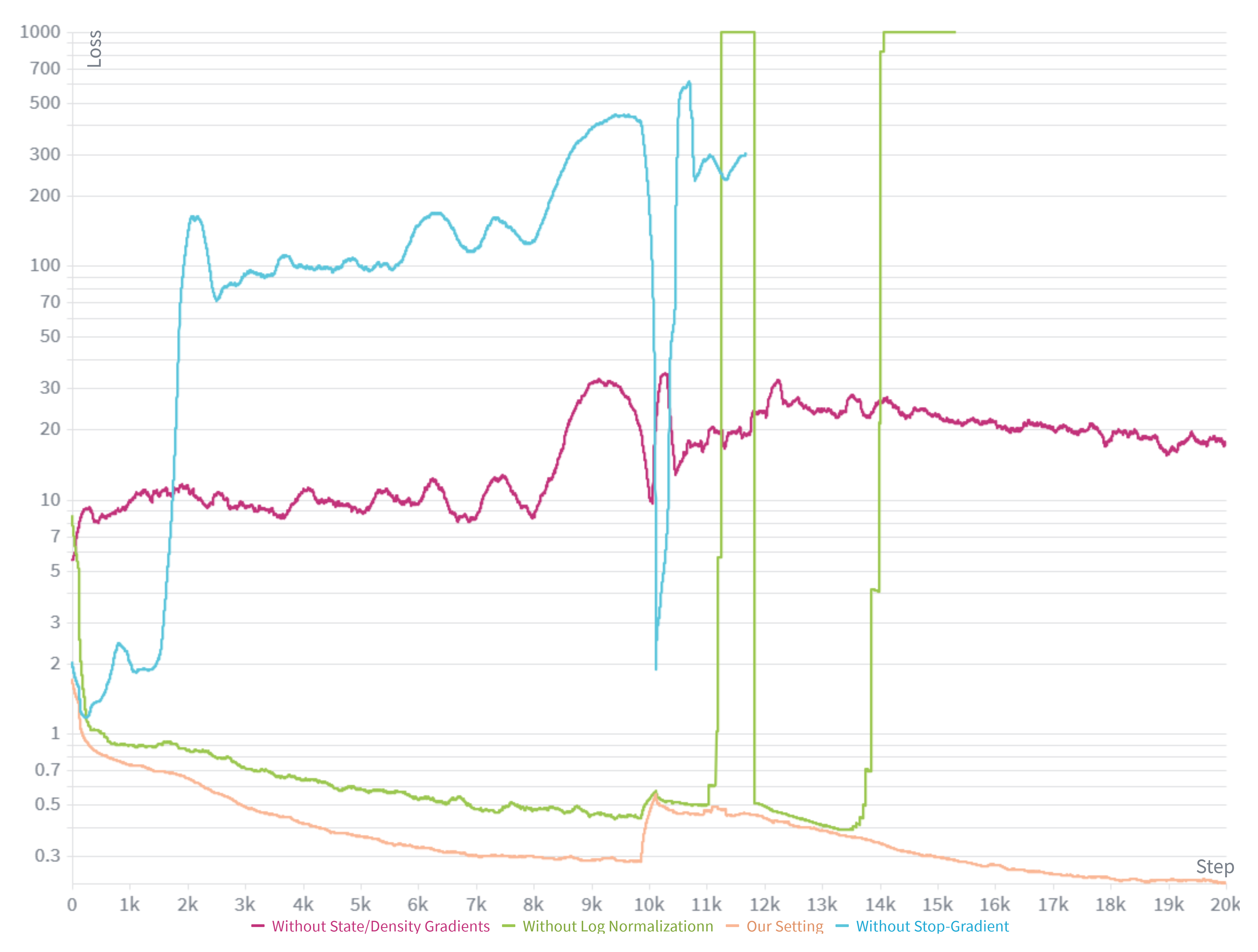}
    \vspace{-8pt}
    \caption{Ablation study on SPH perception: removing gradients prevents learning; removing stop-gradient or log normalization reduces stability and can cause divergence.}
    \vspace{-10pt}
    \label{fig:perception_ablation}
\end{figure}

\begin{figure}[]
    \centering
    \includegraphics[width=\linewidth]{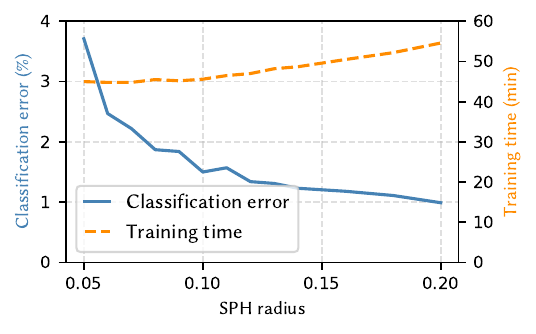}
    \vspace{-15pt}
    \caption{Classification error (left axis) and training time (right axis) vs. SPH radius ($\epsilon$) in PointMNIST classification.}
    \vspace{-5pt}
    \label{fig:mnist-radius}
\end{figure}

We ablate three key components used in our SPH-based perception and training stabilization: (i) stopping gradients through particle positions, (ii) logarithmic normalization of vector-valued perception terms, and (iii) removing directional perception (state gradients and density gradient).
We run all ablations on the 2D morphogenesis setting (lizard target) and report the training loss over iterations.

As shown in \autoref{fig:perception_ablation}, removing directional perception prevents effective learning. Both stabilization choices still allow learning but are substantially less stable: without stopping gradients through positions, training diverges early; without log normalization, training is initially stable but becomes brittle after a pool reset (iteration 10k), often leading to divergence.
Notably, both unstable variants also plateau at higher loss than our full model throughout training, indicating reduced learning capacity in addition to poorer stability.

% \rev{
% We conduct a linear search on SPH radius $\epsilon$ to study its effect on the quality and the runtime of NPA.
% \autoref{fig:mnist-radius} shows the classification accuracy and training time in PointMNIST classification experiment for the range of $\epsilon$, $0.5\times$-$2\times$.
% The figure depicts the trade-off relationship between accuracy and runtime.
% Smaller $\epsilon$ gives noisier perception due to deficient neighbors.
% Larger $\epsilon$ makes NPA run slower, as the time complexity is linearly depends on the average number of neighbors of the particles.
% We use the optimal $\epsilon$ for the experiments in the main text considering this trade-off relationship.
% }

% \begin{figure}[!htbp]
%     \centering
%     \includegraphics[width=\linewidth]{figs/mnist-radius-1.pdf}
%     \vspace{-15pt}
%     \caption{\rev{Classification error and training runtime vs. SPH radius ($\epsilon$) in PointMNIST classification.}}
%     \vspace{-5pt}
%     \label{fig:mnist-radius}
% \end{figure}

We perform a sweep over the SPH support radius $\epsilon$ to study its impact on both accuracy and runtime. 
\autoref{fig:mnist-radius} reports PointMNIST classification error and training time when varying $\epsilon$ over a $0.5\times$--$2\times$ range. 
Smaller $\epsilon$ yields noisier perception due to too few neighbors, degrading accuracy, while larger $\epsilon$ increases the average neighbor count and therefore slows training (runtime scales approximately linearly with the number of neighbors).
We choose $\epsilon$ in the main experiments based on this accuracy-runtime trade-off.

\section{Online Interactive Demo}
\label{sec:appendix_demo}

\begin{figure*}[!htbp]
    \centering
    \includegraphics[width=\linewidth]{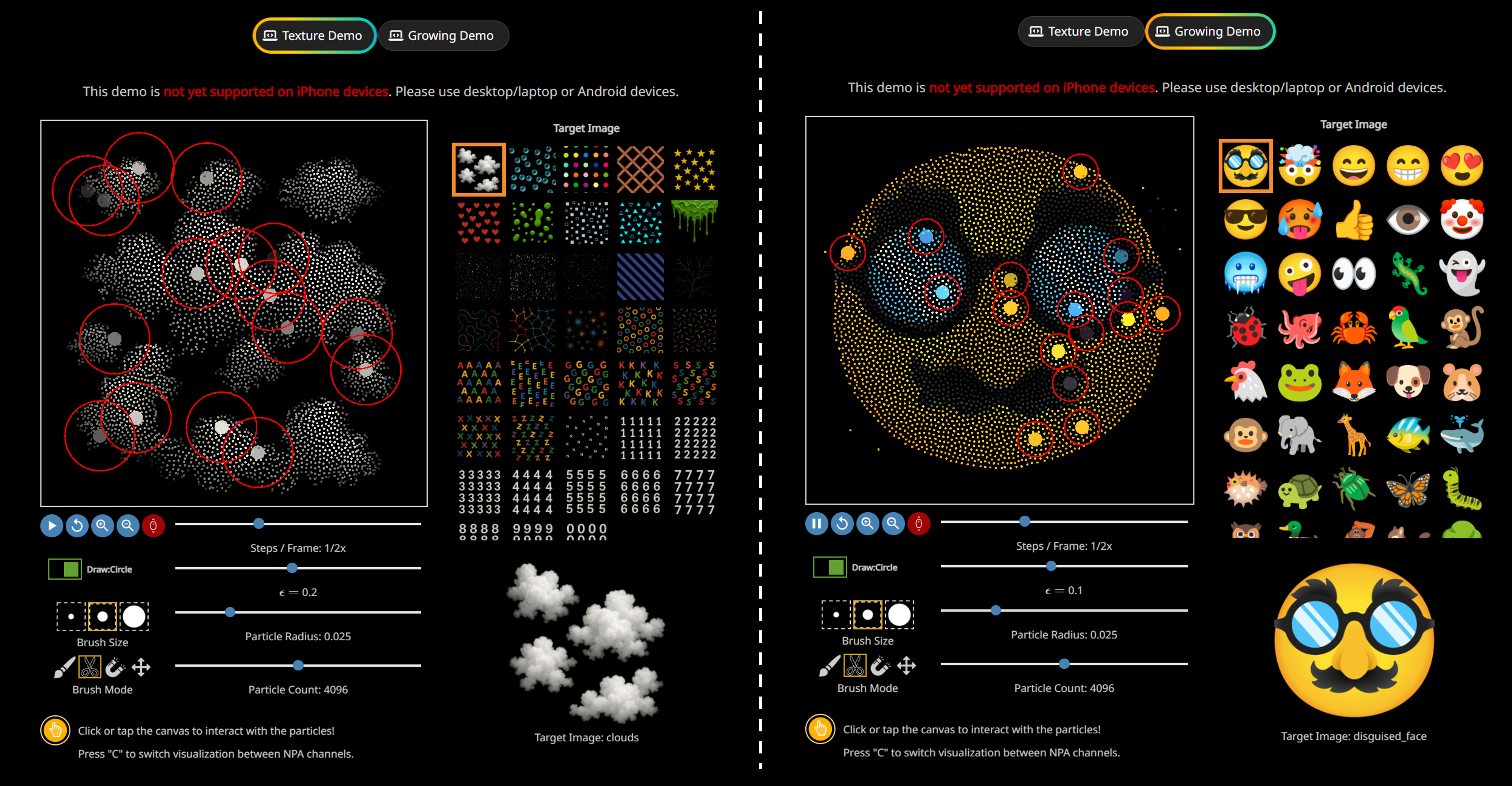}
    \caption{Snapshot of the interactive web demo. Activating trace mode highlights a random set of particles as well as their perception range.}
    \label{fig:demo}
\end{figure*}

To make the learned dynamics tangible in a graphics setting, we provide an online, browser-based interactive demo that runs trained NPA models in real time and allows direct user interaction with the particle system: \textcolor{pinkl}{\href{https://selforg-npa.github.io/}{\textbf{https://selforg-npa.github.io/}}}. The demo comes in two variants---a \emph{Growing Demo} (targets are emoji-like shapes) and a \emph{Texture Demo} (targets are RGBA textures, sourced from DTD \cite{dtd} or generated using a text-to-image generator.). Both variants share the same graphics shader; they differ only in the initialization and the loaded network weights.

Our implementation is built with \emph{SwissGL}, a lightweight wrapper over the WebGL2 API that streamlines the management of GLSL shaders and GPU buffers. We represent the particle set as floating-point textures: each particle stores its multi-channel state and its position, and NPA inference is executed by repeatedly applying a small set of fragment shaders that (i) build an indexing structure for neighborhood queries, (ii) evaluate SPH-style perception and the learned NPA update, and (iii) render particles via Gaussian splatting.

A central challenge for real-time NPA inference in the browser is neighborhood aggregation on a dynamic particle set. To avoid all-pairs interactions, we build a shader-friendly spatial index directly in SwissGL. Particles are sorted by a spatial key using the bitonic sort algorithm. Since particles typically move only slightly per update, we apply a short \emph{maintenance} sort after each step rather than a full re-sorting, which substantially improves performance. After sorting, we compute coarse per-block bounding boxes and use bounding-box intersection tests to identify candidate neighboring blocks, yielding a compact set of blocks to scan during SPH accumulation.

The core NPA computation is implemented as two shaders. First, a density pass estimates (inverse) local density for each particle using the same neighborhood scan. Second, the update pass accumulates SPH perception features (e.g., smoothing and gradients) over the candidate neighborhood and applies the learned adaptation MLP in a fused manner directly within the same shader, with network weights stored as textures. %We introduce a small number of user-facing controls that modulate the simulation without retraining.

For visualization, each particle is rendered as a Gaussian splat or a circle into an image buffer. We use a test-time compositing rule $d\cdot(1-s_a) + s$
where $d$ is the current buffer value, $s$ is the fragment-emitted color and $s_a$ its alpha. This blending prevents color saturation when many splats overlap and produces stable, readable renderings during interaction. The demo also includes a \emph{trace} option that highlights a sparse subset of particles by rendering them with larger splats, making long-term trajectories and persistent flows easier to observe. The trace mode also shows the the perception radius of the traced particles, $\epsilon$, as a red circle around them, as shown in \autoref{fig:demo}.

The UI exposes interactive controls commonly needed to explore NPA behavior: support radius $\epsilon$, simulation speed (steps per frame), splat radius, and particle count. Particle count is restricted to powers of two to match the bitonic sort implementation. We also provide brush-based perturbations to demonstrate robustness and regeneration under user intervention:
(i) zeroing particle states in a region, (ii) pushing particles along a stroke (cut/scissor-like), and (iii) pulling particles toward a point (magnet-like). These tools let users directly probe the locality, stability, and regenerative properties of the learned particle dynamics in a way that complements the offline figures and videos.

\end{document}